\documentclass[journal]{IEEEtran}
\usepackage[utf8]{inputenc}
\usepackage[table]{xcolor}

\usepackage{fixltx2e}

\usepackage{graphicx}

\usepackage{subcaption}              %subfigures with \mbox and \subfigure[]

\usepackage[utf8]{inputenc}

\usepackage{amsfonts}
\usepackage{amsmath}
\usepackage{mathtools}
\usepackage{amssymb}

\usepackage{bm}

\usepackage{color,verbatim}
\usepackage{multirow}
\usepackage{accents}

\usepackage{theoremref}
\usepackage{flushend}

\usepackage{url}

\newcounter{rulecounter}
\newcommand{\resetrule}{ \setcounter{rulecounter}{0}}
\resetrule

\newsavebox{\selvestebox}
\newenvironment{colbox}[1]
  {\newcommand\colboxcolor{#1}%
   \begin{lrbox}{\selvestebox}%
   \begin{minipage}{\dimexpr\columnwidth-2\fboxsep\relax}}
  {\end{minipage}\end{lrbox}%
   \begin{center}
   \colorbox{\colboxcolor}{\usebox{\selvestebox}}
   \end{center}}

\definecolor{orange}{rgb}{1,0.8,0}
\definecolor{gray}{rgb}{.9,0.9,0.9}
\definecolor{darkgray}{rgb}{.3,0.3,0.3}
\definecolor{darkblue}{rgb}{.1,0.0,0.3}
\definecolor{lightblue}{rgb}{0.7,0.7,1}
\definecolor{lightred}{rgb}{1,0.7,.7}
\definecolor{purple}{RGB}{204,153,255}
\definecolor{lightgray}{rgb}{.95,0.95,0.95}
\definecolor{lightgreen}{rgb}{0.3,0.5,0.3}
\definecolor{darkgreen}{rgb}{0.05,0.3,0.05}

\newcommand{\inv}{^{-1}}

\newcommand{\rfield}{\mathbb{R}}

\newtheorem{myproposition}{Proposition}
\newtheorem{myremark}{Remark}
\newtheorem{myproblemstatement}{Problem Statement}
\newtheorem{mylemma}{Lemma}
\newtheorem{mytheorem}{Theorem}
\newtheorem{mydefinition}{Definition}
\newtheorem{mycorollary}{Corollary}

\usepackage{pgfplots}
\pgfplotsset{compat=newest}
\usetikzlibrary{plotmarks}
\usetikzlibrary{positioning}
\usetikzlibrary{arrows.meta}
\usepgfplotslibrary{patchplots}
\usepackage{grffile}

\pgfplotsset{plot coordinates/math parser=false}
\newlength\mywidth
\newlength\myheight
\newlength\mywidths
\newlength\myheights
\newlength\mywidthss
\newlength\myheightss
\definecolor{mycolor1}{rgb}{0.00000,0.44700,0.74100}%
\definecolor{mycolor2}{rgb}{0.85000,0.32500,0.9800}%
\definecolor{mycolor3}{rgb}{0.92900,0.69400,0.12500}%
\definecolor{mycolor4}{rgb}{0.89400,0.18400,0.15600}%
\definecolor{mycolor5}{rgb}{0.46600,0.67400,0.18800}%
\definecolor{mycolor6}{rgb}{0.30100,0.74500,0.93300}%
\definecolor{mycolor7}{rgb}{0.63500,0.07800,0.18400}%

\definecolor{AGNN}{rgb}{1,0.04700,0.04100}%
\definecolor{GCN}{rgb}{0.05000,0.32500,0.89800}%
\definecolor{AGNNnf}{rgb}{1,0.8,0.119800}%
\definecolor{Mune}{rgb}{0.2,0.800,1}%

\definecolor{TightHann}{rgb}{0.92900,0.69400,0.12500}%
\definecolor{pDiffScater}{rgb}{0.49400,0.18400,0.55600}%
\definecolor{pMonicCubic}{rgb}{0.46600,0.67400,0.18800}%
\definecolor{pTightHann}{rgb}{0.30100,0.74500,0.93300}%

\pgfplotscreateplotcyclelist{colorlist}{%
color=AGNN,line width=2.0pt,
solid, every mark/.append style={solid, fill=AGNN}, mark=*\\%
color=GCN,line width=2.0pt,dotted, every mark/.append style={solid, fill=GCN}, mark=square*\\%
color=AGNNnf,line width=2.0pt,densely dotted, every mark/.append style={solid, fill=AGNNnf}, mark=otimes*\\%
color=Mune,line width=2.0pt,loosely dotted, every mark/.append style={solid, fill=Mune}, mark=triangle*\\%
color=mycolor5,line width=2.0pt,
dashed, every mark/.append style={solid, fill=mycolor5},mark=diamond*\\%
color=mycolor6,line width=2.0pt,loosely dashed, every mark/.append style={solid, fill=mycolor6},mark=*\\%
color=mycolor7,densely dashed, every mark/.append style={solid, fill=mycolor7},line width=2.0pt,mark=square*\\%
dashdotted, every mark/.append style={solid, fill=mycolor7},mark=otimes*\\%
dashdotdotted, every mark/.append style={solid},mark=star\\%
densely dashdotted,every mark/.append style={solid, fill=gray},mark=diamond*\\%
}
\pgfplotscreateplotcyclelist{my black white}{%
solid, every mark/.append style={solid, fill=gray}, mark=*\\%
dotted, every mark/.append style={solid, fill=gray}, mark=square*\\%
densely dotted, every mark/.append style={solid, fill=gray}, mark=otimes*\\%
loosely dotted, every mark/.append style={solid, fill=gray}, mark=triangle*\\%
dashed, every mark/.append style={solid, fill=gray},mark=diamond*\\%
loosely dashed, every mark/.append style={solid, fill=gray},mark=*\\%
densely dashed, every mark/.append style={solid, fill=gray},mark=square*\\%
dashdotted, every mark/.append style={solid, fill=gray},mark=otimes*\\%
dashdotdotted, every mark/.append style={solid},mark=star\\%
densely dashdotted,every mark/.append style={solid, fill=gray},mark=diamond*\\%
}
\relax
\usetikzlibrary{decorations.pathreplacing}
\newcommand{\xlabelfontsize}{\small}

\newcommand{\legendfontsize}{\small}
\newcommand{\ticklabelfontsize}{\small}
\newcommand{\mlabelfontsize}{\small}
\definecolor{lavander}{cmyk}{0,0.48,0,0}
\definecolor{violet}{cmyk}{0.79,0.88,0,0}
\definecolor{burntorange}{cmyk}{0,0.52,1,0}
\definecolor{mygreen}{rgb}{0,1,0}%
\definecolor{myred}{rgb}{1,0,0}%

\newcommand{\mylinewidth}{1.5}
\newcommand{\markwidth}{1.5}
\newcommand{\distbetweenlayers}{1.8}
\tikzstyle{nnblock}=[draw,rectangle,  rounded corners,text=black,
minimum width=30pt,minimum height=60pt]
\tikzstyle{esl}=[draw,rectangle,  rounded corners,text=black,
minimum width=60pt,minimum height=30pt]
\tikzstyle{grnn}=[draw,rectangle,  rounded corners,text=black,
minimum width=30pt,minimum height=15pt]
\tikzstyle{graphsampler}=[draw,rectangle,  fill=orange,fill opacity=0.1, rounded corners,text=orange,
minimum width=30pt,minimum height=15pt]
\tikzstyle{outblock}=[draw,rectangle, color=red, rounded corners,text=black,
minimum width=10pt,minimum height=60pt]
\tikzstyle{nam}=[rectangle, fill=orange,fill opacity=0.1, rounded corners,text=orange,
minimum width=8pt,minimum height=58pt]
\tikzstyle{gam}=[fill=blue,fill opacity=0.1, rectangle,  rounded corners,text=black,
minimum width=8pt,minimum height=58pt]
\tikzstyle{fam}=[fill=green,fill opacity=0.1, rectangle,  rounded corners,text=black,
minimum width=8pt,minimum height=58pt]

\tikzstyle{peers}=[draw,circle,bottom color=myred,
                  top color= white, text=violet,minimum width=8pt]
\tikzstyle{superpeers}=[draw,circle, left color=mygreen,
                       text=violet,minimum width=8pt]

\tikzstyle{peers3}=[draw,circle,bottom color=myred,%lavander,
                  top color= white, text=violet,minimum width=12pt]
\tikzstyle{superpeers3}=[draw,circle, left color=mygreen,%orange,
                       text=violet,minimum width=12pt]
\tikzstyle{peers2}=[draw,circle,bottom color=myred,
                  top color= white, text=violet,inner sep=0pt,minimum size=0.1cm]
\tikzstyle{superpeers2}=[draw,circle, left color=mygreen,
                       text=violet,inner sep=0pt,minimum size=0.1cm]
                    
\tikzstyle{peers1}=[draw,circle,bottom color=myred,%lavander,
                  top color= white, text=violet,minimum width=4pt]
\tikzstyle{superpeers1}=[draw,circle, left color=mygreen,%orange,
                       text=violet,minimum width=4pt]
\tikzstyle{legendsp}=[rectangle, draw,  rounded corners,
                     thin,bottom color=mygreen, top color=white,
                     text=black, minimum width=2cm]
\tikzstyle{legendp}=[rectangle, draw,  rounded corners, thin,
                     bottom color=myred, top color= white,
                     text= black, minimum width= 2cm]
\tikzstyle{legend_general}=[rectangle, rounded corners, thin,
                           burntorange, fill= white, draw, text=violet,
                           minimum width=2.5cm, minimum height=0.8cm]
\usetikzlibrary{shapes.misc, positioning}

\usetikzlibrary{calc,trees}

\newcommand{\convexpath}[2]{
[   
    create hullnodes/.code={
        \global\edef\namelist{#1}
        \foreach [count=\counter] \nodename in \namelist {
            \global\edef\numberofnodes{\counter}
            \node at (\nodename) [draw=none,name=hullnode\counter] {};
        }
        \node at (hullnode\numberofnodes) [name=hullnode0,draw=none] {};
        \pgfmathtruncatemacro\lastnumber{\numberofnodes+1}
        \node at (hullnode1) [name=hullnode\lastnumber,draw=none] {};
    },
    create hullnodes
]
($(hullnode1)!#2!-90:(hullnode0)$)
\foreach [
    evaluate=\currentnode as \previousnode using \currentnode-1,
    evaluate=\currentnode as \nextnode using \currentnode+1
    ] \currentnode in {1,...,\numberofnodes} {
-- ($(hullnode\currentnode)!#2!-90:(hullnode\previousnode)$)
  let \p1 = ($(hullnode\currentnode)!#2!-90:(hullnode\previousnode) - (hullnode\currentnode)$),
    \n1 = {atan2(\y1,\x1)},
    \p2 = ($(hullnode\currentnode)!#2!90:(hullnode\nextnode) - (hullnode\currentnode)$),
    \n2 = {atan2(\y2,\x2)},
    \n{delta} = {-Mod(\n1-\n2,360)}
  in 
    {arc [start angle=\n1, delta angle=\n{delta},radius=#2]}
}
-- cycle
}
\newcommand{\featvec}{\mathbf{x}}
\newcommand{\dataentry}{X}
\newcommand{\datamatrix}{\mathbf{\dataentry}}

\newcommand{\nbrnodes}{N}

\newcommand{\shiftind}{i}
\newcommand{\nbrshifts}{I}

\newcommand{\nbrlayers}{L}

\newcommand{\graph}{\mathcal{G}}

\newcommand{\nodeshiftnot}[2]{_{#1#2}}

\newcommand{\layernot}[1]{^{(#1)}}
\newcommand{\layerind}{l}

\newcommand{\weightmat}{\mathbf{W}}

\newcommand{\nodeind}{n}

\newcommand{\weighttensor}{\underline{\weightmat}}

\newcommand{\outputlaymat}{\check{\mathbf{Z}}}

\newcommand{\outputlaytensor}{\underline{\outputlaymat}}

\newcommand{\diffusedfeatvec}{\mathbf{h}}
\newcommand{\graphadaptivec}{\mathbf{g}}

\newcommand{\nbrclasses}{K}

\newcommand{\predictionmat}{\hat{\mathbf{Y}}}

\newcommand{\regfun}{\rho}

\usepackage{array}
\usepackage{footnote}
\makesavenoteenv{tabular}
\makesavenoteenv{table}

\usepackage{booktabs}       % professional-quality tables

\usepackage{floatrow}
\newfloatcommand{capbtabbox}{table}[][\FBwidth]

\title{Tensor Graph Convolutional Networks for Multi-relational and Robust Learning}
\author{Vassilis~N.~Ioannidis,~\IEEEmembership{Student Member,~IEEE,}
        Antonio~G.~Marques,~\IEEEmembership{Senior Member,~IEEE,}
         Georgios~B.~Giannakis,~\IEEEmembership{Fellow,~IEEE}
\thanks{The work in this paper has been supported by the Doctoral Dissertation Fellowship of the Univ. of Minnesota, the USA NSF grants 171141,  1500713, and  1442686, and by the Spanish grants MINECO KLINILYCS (TEC2016-75361-R) and Instituto de Salud Carlos III DTS17/00158.}
\thanks{V. N. Ioannidis and G. B. Giannakis are with the Dept. of Electrical and Computer Engineering,
  Univ. of Minnesota,
  Minneapolis, MN, USA (e-mail:, ioann006@umn.edu;
georgios@umn.edu).}% <-this % stops a space
\thanks{A. G. Marques is with the Dept. of Signal Theory and Communications, King Juan Carlos Univ., Madrid, Spain (e-mail:, antonio.garcia.marques@urjc.es).}% <-this % stops a space
}

\begin{document}

\maketitle
\begin{abstract}
The era of ``data deluge'' has sparked renewed interest in graph-based learning methods and their widespread applications ranging from sociology and biology to transportation and communications. In this context of graph-aware methods, the present paper introduces a tensor-graph convolutional network (TGCN) for scalable semi-supervised learning (SSL) from data associated with a collection of graphs, that are represented by a tensor. Key aspects of the novel TGCN architecture are the dynamic adaptation to different relations in the tensor graph via learnable weights, and the consideration of graph-based regularizers to promote smoothness and alleviate over-parameterization. The ultimate goal is to design a powerful learning architecture able to: discover complex and highly nonlinear data associations, combine (and select) multiple types of relations, scale gracefully with the graph size, and remain robust to perturbations on the graph edges. The proposed architecture is relevant not only in applications where the nodes are naturally involved in different relations (e.g., a multi-relational graph capturing family, friendship and work relations in a social network), but also in robust learning setups where the graph entails a certain level of uncertainty, and the different tensor slabs correspond to different versions (realizations) of the nominal graph.  Numerical tests showcase that the proposed architecture achieves markedly improved performance relative to standard GCNs, copes with state-of-the-art adversarial attacks, and leads to remarkable  SSL performance over protein-to-protein interaction networks.
\end{abstract}
\section{Introduction}
A task of major importance at the interface of machine learning with network science is semi-supervised learning (SSL) over graphs. In a nutshell, SSL aims at predicting or extrapolating nodal attributes given: i) the values of those attributes at a subset of nodes and (possibly) ii) additional features at all nodes. A relevant example is protein-to-protein interaction networks,  where the proteins (nodes) are associated with specific biological functions (the nodal attributes in this case are binary values indicating whether the protein participates in the function or not), thereby facilitating the understanding of pathogenic and physiological mechanisms.

While significant progress has been made, most works consider that the relation among nodal variables is represented by a single graph. This may be inadequate in many contemporary applications, where nodes may engage in multiple types of relations~\cite{kivela2014multilayer}, motivating the generalization of traditional SSL approaches to \emph{single-relational} graphs to \emph{multi-relational} (a.k.a. multi-layer) graphs.
%\footnote{{Many works refer to these graphs as multi-layer graphs~\cite{kivela2014multilayer}.}}. 
In the particular case of protein interaction networks, each layer of the graph could correspond to a different type of tissue, e.g., brain or muscle. In  a social network, each layer could amount to a form of social interaction, such as friendship, family bonds, or coworker-ties \cite{wasserman1994social}. Such graphs can be represented by a tensor graph, where each tensor slab corresponds to a single relation. With their ubiquitous presence granted, {the} development of SSL methods that account for multi-relational networks is only in its infancy, see, e.g.,~\cite{kivela2014multilayer,ioannidis2018multilay}. This work develops a novel \emph{robust} deep learning framework for SSL over \emph{multi-relational} graphs. 

Graph-based SSL methods typically assume that the true labels are ``smooth'' over the graph,  which naturally motivates leveraging the network topology to propagate the labels and increase learning performance.  Graph-induced smoothness can be captured by graph kernels~\cite{belkin2006manifold,smola2003kernels,ioannidis2018kernellearn};  Gaussian random fields \cite{zhu2003semi}; or  low-rank {parametric} models~\cite{shuman2013emerging}. Alternative approaches use the graph to embed nodes in a vector space, and then apply learning approaches to the resultant vectors~\cite{weston2012deep,yang2016revisiting,berberidis2018adaptive}. More recently, the map from input data to their labels is given by a neural network (NN) that incorporates the graph structure and generalizes the typical convolution operations; see e.g., \cite{bronstein2017geometric,gama2018convolutionaljournal,ioannidis2018graphrnn,schlichtkrull2018modeling}. The parameters describing the graph convolutional NN (GCN) are then learned using labeled examples and feature vectors, and those parameters are employed to predict labels of the unobserved nodes; see, e.g., \cite{kipf2016semi,bronstein2017geometric,velivckovic2017graph}, for state-of-the-art results in SSL when nodes are attributed with features. 
	
With the success of GCNs on graph learning tasks granted, recent reports point out that perturbations of the graph topology can severely deteriorate learning performance \cite{zugner18adv,xu2019topology,dai2018adversarial}. Such uncertainty in the topology may be attributed to several reasons. First, the graph is implicit and its topology is identified using data-driven methods~\cite{giannakis2017tutor}. However, each method relies on a different model and assumptions, and without ground truth selecting the appropriate graph-learning technique is challenging. A less accurate model can induce perturbations to the learned graph. Further, in random graph models one deals with a realization of {a} graph whose edges may be randomly perturbed~\cite{newman2018networks}. Similarly, this is also relevant in adversarial settings, where the links of the nominal graph are corrupted by some foe that aims to poison the learning process. Adversarial perturbations target a subset of nodes and modify their links to promote {the} miss-classification of targeted nodes~\cite{wu19adv}. Crafted graph perturbations are ``unnoticeable,'' which is feasible so long as the degree distribution of the perturbed graphs is similar to the initial distribution~\cite{zugner18adv}.  GCNs learn nodal representations by extracting information within local neighborhoods. These learned features may be significantly perturbed if the neighborhood is altered. Hence, this vulnerability of GCNs challenges their deployment in critical applications dealing with security or healthcare, where robust learning is of paramount importance. Defending against adversarial, random, or model-based perturbations may unleash the potential of GCNs, and broaden the scope of machine learning applications altogether. 

\vspace{.1cm}
\noindent\textbf{Contributions.} 
This paper develops a deep SSL approach over multiple graphs with applications to both multi-relational data and robust learning. Specifically, the contribution is five-fold.
\begin{itemize}
\item[\textbf{C1}.] A tensor-based GCN is developed to account for multi-relational graphs. Learnable coefficients are introduced to effect model adaptivity to multiple graphs, and identification of the underlying data structure.
\item[\textbf{C2}.] A multi-hop convolution is introduced along with a residual data feed per graph, thus broadening the class of (graph signal) transformations the GCN implements; and hence, facilitating the diffusion of nodal features across the graph. In the training phase suitable (graph-based) regularizers are incorporated to guard against overfitting, and further capitalize on the graph topology. 
\item[\textbf{C3}.] For nodes involved in different relations, and for (multi-relational) datasets adhering to several graphs, the proposed TGCN provides a powerful SSL approach by leveraging the information codified across multiple graphs.
\item[\textbf{C4}.] The novel TGCN enables robust SSL for single- or multi-relational data when the underlying topology is perturbed. Perturbations include model induced, random, and adversarial ones.  To defend against adversaries, a novel edge-dithering (ED) approach is developed that generates ED graphs by sampling edges of the original graph with probabilities selected to enhance robustness.
\item[\textbf{C5}.] Numerical tests with multi-relational protein networks showcase the merits of the proposed tensor-graph framework. Further experiments with noisy features, noisy edge weights, and random as well as adversarial edge perturbations verify the robustness of our novel approach.
\end{itemize}

\noindent\textbf{Notation.} Scalars are denoted by lowercase, column vectors by bold lowercase, matrices  by bold uppercase, and tensors using bold uppercase underscored letters. Superscripts $~^{\top}$ and $~\inv$ denote, respectively, the transpose and inverse operators; while $\boldsymbol 1_N$ stands for the $N\times1$ all-one  vector. Finally, if $\mathbf A$ is a nonsingular matrix and $\mathbf x$ a vector, then $||\mathbf x||^2_{\mathbf A}:= \mathbf x^{\top} \mathbf A\inv \mathbf x$, $\|\mathbf A\|_1$ denotes  the $\ell_1$-norm of the vectorized matrix, and $\|\mathbf A\|_F$ is the Frobenius norm of $\mathbf A$.

\section{SSL over multi-relational graphs}\label{sec:probform}
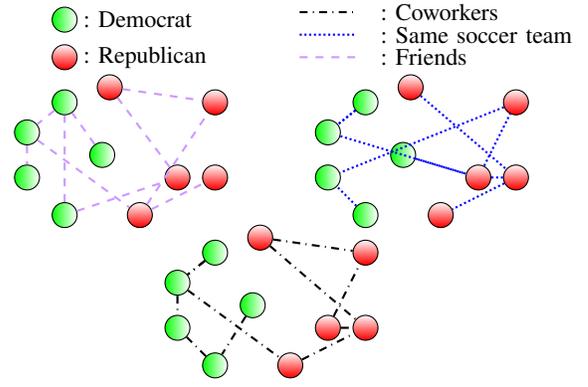
\begin{figure}
    \centering
    \begin{tikzpicture}[%
 block/.style={rectangle, fill=gray, text centered, minimum height=8mm, rounded corners=1mm}
        ,my line style1/.style={ draw=purple,thick}
        ,my line style2/.style={ draw=blue,thick}
        ,my line style3/.style={ draw=black,dashdotted,thick}
        ,every label/.append style = {label distance=1pt, inner sep=1pt, 
                             align=left, font=\tiny}
        %,every edge/.style={my line style}
        ,my edge/.style={my line style, mylabel=#1}
        ,my curved edge/.style n args={4}{my line style, out=#1, in=#2, looseness=#3, mylabel=#4}
        ,edge1/.style={my line style1,dashed}
        ,mylabel1/.style={my label style{#1}}
        ,edge2/.style={my line style2, densely dotted}
        ,edge3/.style={my line style3}
        ,mylabel/.style={edge node={node[my label style]{#1}}}]
  % Place super peers and connect them
\foreach \distxbetweengraphs\distybetweengraphs\graphid in {{0/2/0},{4/2/1},{2/0/2}}
  \foreach \place/\name in {{(0+\distxbetweengraphs,-0.5+\distybetweengraphs)/ar\graphid}, {(0.5+\distxbetweengraphs,0.3+\distybetweengraphs)/br\graphid}, {(0+\distxbetweengraphs,1+\distybetweengraphs)/cr\graphid}, {(-0.5+\distxbetweengraphs,0.6+\distybetweengraphs)/dr\graphid},
           {(-0.5+\distxbetweengraphs,0+\distybetweengraphs)/er\graphid}}
    \node[superpeers] (\name) at \place {};
  
   %
   % Place normal peers
  \foreach \distxbetweengraphs\distybetweengraphs\graphid in {{0/2/0},{4/2/1},{2/0/2}}
  \foreach\place/\name  in {{(1+\distxbetweengraphs,-0.5+\distybetweengraphs)/ad\graphid}, {(1.5+\distxbetweengraphs,0+\distybetweengraphs)/bd\graphid}, {(0.6+\distxbetweengraphs,1.2+\distybetweengraphs)/cd\graphid}, {(2+\distxbetweengraphs,1+\distybetweengraphs)/dd\graphid},
                        {(2+\distxbetweengraphs,0+\distybetweengraphs)/ed\graphid}}
    \node[peers] (\name) at \place {};

   \foreach \source/\dest in {ar0/bd0, ar0/cr0, ad0/dd0,bd0/cd0,  br0/cr0, cr0/dr0,ad0/ed0,dr0/er0,
   cd0/dd0,dr0/ad0}
    \path (\source) edge[edge1] (\dest);
    \foreach \source/\dest in {ar1/er1,bd1/dd1,cd1/ed1, cr1/dr1,bd1/ed1,dd1/er1,
   cr1/dr1,dr1/bd1,br1/bd1,ed1/ad1}
    \path (\source) edge[edge2] (\dest);
    \foreach \source/\dest in {ar2/er2,bd2/dd2,cd2/ed2, cr2/dr2,bd2/ed2, cr2/dr2,ad2/ed2,dr2/er2,
   cd2/dd2,dr2/ad2,br2/ar2}
    \path (\source) edge[edge3] (\dest);
   %%%%%%%%
   % Legends
    \node[superpeers,label={[font=\small]0: : Democrat}] (d) at (0,4.1) {};
    \node[peers,label={[font=\small]0: : Republican}] (r) at (0,3.6) {};
   %\node[legendsp] at  {\small{Democrats}};
   %\node[legendp] at  {\small{Rebublicans}};
   %\node[legend_general] at (0,4) {\small{\textsc{Skype-topology}}};
    \node at (3,3.6) (nA) {};
    \node[label={[font=\small]0: : Friends}] at (4,3.6) (nB) {};
    \path (nA) edge[edge1] (nB);
    \node at (3,3.9) (nA1) {};
    \node[label={[font=\small]0: : Same soccer team}] at (4,3.9) (nB1) {};
    \path (nA1) edge[edge2] (nB1);
    \node at (3,4.2) (nA2) {};
    \node[label={[font=\small]0: : Coworkers}] at (4,4.2) (nB2) {};
    \path (nA2) edge[edge3] (nB2);
\end{tikzpicture}
    \caption{A multi-relational network of voters.}
    \label{fig:multilayer}
\end{figure}

Consider a network of ${N}$ nodes, with nodal (vertex) set $\mathcal{V}:=\{v_1,\ldots,v_{N}\}$, connected through ${I}$ relations. The ${i}$th relation is captured by  the  ${N}\times{N}$ adjacency matrix $\mathbf{A}_{i}$, whose entry $A_{nn'i}$ represents the weight of the edge connecting nodes  $v_{n}$ and $v_{n'}$ as effected by the ${i}$th relation. The matrices $\{\mathbf{A}_{i}\}_{{i}=1}^{I}$ are collected in the ${N}\times{N}\times{I}$ tensor $\underline{\mathbf{A}}$. In the social network examples already provided in the previous section, each $i$ could for instance represent a relation via a particular  app, such as Facebook, LinkedIn, or Twitter; see Fig.\ref{fig:multilayer}. Regardless of the application, the neighborhood of $v_{n}$ induced by relation ${i}$ is specified by the set
\begin{align}
\label{eq:neighborhood}
\mathcal{N}_{n}^{({i})}:=\{{n'}:A_{nn'i}\ne0,~~ v_{n'}\in\mathcal{V}\}.
\end{align}

We further associate an $ F \times 1$ feature vector $\mathbf{x}_{{n}}$ with the ${n}$th node, and collect those vectors in the $N \times F$ feature matrix $\mathbf{X} :=[\mathbf{x}_{1},\ldots,\mathbf{x}_{{N}}]^\top$, where entry $X_{{n}p}$ may denote e.g., the salary of individual $n$ in the LinkedIn social network. 

Each node ${n}$ has a label $y_{n}\in\{0,\ldots,K-1\}$, which in the last example could represent the education level of a person. In SSL, we know labels only for a subset of nodes $\{y_{{n}}\}_{{n}\in\mathcal{M}}$, with $\mathcal{M} \subset\mathcal{V}$. This partial availability may be attributed to privacy concerns (medical data); energy considerations  (sensor networks); or unrated items (recommender systems). The ${N}\times K$ matrix  $\mathbf{Y}$ is the ``one-hot'' representation of the true nodal labels; that is, if $y_{n}=k$ then $Y_{{n},k}=1$ and $Y_{{n},k'}=0, \forall k'\ne k$.

Given $\mathbf{X}$ and $\underline{\mathbf{A}}$, the goal is to develop a \textit{robust tensor-based deep} SSL approach over \textit{multi-relational} graphs; that is, develop a TGCN mapping each node $n$ to its label $y_{n}$; and hence, learn the unavailable labels. 

\section{Proposed TGCN architecture}
Deep learning architectures typically process the input information using a succession of $L$ hidden layers. Each of the layers comprises a conveniently parametrized linear transformation, a scalar nonlinear transformation, and possibly a dimensionality reduction (pooling) operator. By successively combining (non)linearly local features, the aim at a high level is to progressively extract useful information for learning~\cite{goodfellow2016deep}. GCNs tailor these operations to the graph that supports the data~\cite{bronstein2017geometric}, including the linear \cite{defferrard2016convolutional}, and nonlinear \cite{defferrard2016convolutional} operators.  
In this section, we describe the blocks of our novel multi-relational TGCN, which inputs the known features at the first layer, and outputs the predicted labels at the last layer. We first present the TGCN operation, the output layers, and finally discuss how training is performed.

\subsection{Single layer operation}
Consider the output ${N}\times{I}\times P^{(l)}$ tensor  $\check{\underline{\mathbf{Z}}}^{(l)}$ of an intermediate layer, say the $l$th one, that holds the $P^{(l)}\times 1$  feature vectors $\check{\mathbf{z}}_{{n}{i}}^{(l)}, \forall {n},{i}$, with $P^{(l)}$ being the number of output features at $l$. Similarly, the  ${N}\times{I}\times P^{(l-1)}$ tensor $\check{\underline{\mathbf{Z}}}^{(l-1)}$ represents the input of layer $l$. 
%Since our focus is on predicting labels of all nodes, we do not consider a dimensionality reduction (pooling) operator in the intermediate layers. 
The mapping from $\check{\underline{\mathbf{Z}}}^{(l-1)}$ to $\check{\underline{\mathbf{Z}}}^{(l)}$ consists of two sub-maps. A linear one that maps the  ${N}\times{I}\times P^{(l)}$ tensor $\check{\underline{\mathbf{Z}}}^{(l-1)}$ to the ${N}\times{I}\times P^{(l)}$ tensor $\underline{\mathbf{Z}}^{(l)}$; followed by a memoryless scalar nonlinearity $\sigma({\cdot})$ applied to $\underline{\mathbf{Z}}^{(l)}$  as 
\begin{align}\label{eq:nonlinear}
\check{\underline{{Z}}}_{{i}{n}p}^{(l)}:=\sigma(
{\underline{{Z}}_{{i}{n}p}^{(l)}}).
\end{align}
The output $\check{\underline{\mathbf{Z}}}^{(l)}$ of layer $l$ is formed by the entries in 
\eqref{eq:nonlinear}. A common choice for $\sigma{(\cdot)}$ is the rectified linear unit (ReLU), that is, $\sigma{(c)}=\text{max}(0,c)$ \cite{goodfellow2016deep}. The linear map from  $\check{\underline{\mathbf{Z}}}^{(l-1)}$ to $ \underline{\mathbf{Z}}^{(l)}$ will be designed during training. Convolutional NNs (CNNs) typically consider a small number of trainable weights and then generate the linear output as a convolution of the input with these weights~\cite{goodfellow2016deep}. The convolution combines values of close-by inputs (consecutive time instants, or neighboring pixels) and thus extracts information of local neighborhoods. Permeating CNN benefits to the graph domain, GCNs replace the convolution with a `graph filter' whose taps are learnable~\cite{bronstein2017geometric}. Graph filters can have low order (degrees of freedom), and certainly account for the graph structure.

In the following three subsections, we introduce the structure of the novel tensor-graph linear transformation, and elaborate on how the multi-relational graph is taken into account. 

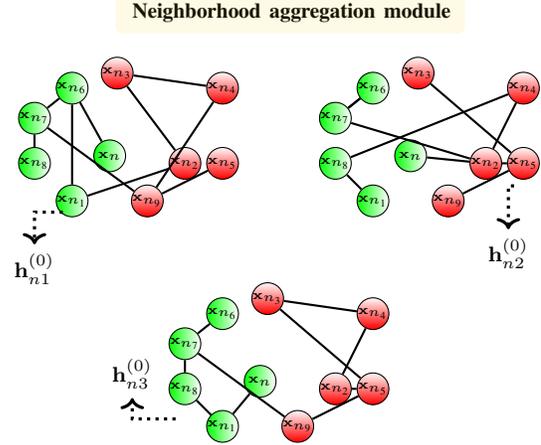
\begin{figure}
    \centering
    \begin{tikzpicture}[%
 block/.style={rectangle, fill=gray, text centered, minimum height=8mm, rounded corners=1mm}
        ,my line style1/.style={ draw=purple,dashed,thick}
        ,my line style2/.style={ draw=blue,densely dotted,thick}
        ,my line style3/.style={ draw=black,dashdotted,thick}
        ,my line style5/.style={ draw=black,thick}
        ,my line style4/.style={ draw=orange,dotted, very thick}
        ,my line style6/.style={ draw=black,thick}
        ,every label/.append style = {label distance=1pt, inner sep=1pt, 
                             align=left, font=\tiny}
        %,every edge/.style={my line style}
        ,my edge/.style={my line style, mylabel=#1}
        ,my curved edge/.style n args={4}{my line style, out=#1, in=#2, looseness=#3, mylabel=#4}
        ,edge1/.style={my line style5}
        ,mylabel1/.style={my label style{#1}}
        ,edge2/.style={my line style5}
        ,edge3/.style={my line style5}
         ,edge4/.style={my line style6}
        ,mylabel/.style={edge node={node[my label style]{#1}}}]
  % Place super peers and connect them
\foreach \distxbetweengraphs\distybetweengraphs\graphid in {{0/3/0},{4/3/1},{2/0/2}}
  \foreach \place/\name in {{(0+\distxbetweengraphs,-0.5+\distybetweengraphs)/ar\graphid}, {(0.5+\distxbetweengraphs,0.1+\distybetweengraphs)/br\graphid}, {(0+\distxbetweengraphs,1+\distybetweengraphs)/cr\graphid}, {(-0.5+\distxbetweengraphs,0.6+\distybetweengraphs)/dr\graphid},
           {(-0.5+\distxbetweengraphs,0+\distybetweengraphs)/er\graphid}}
    \node[superpeers3] (\name) at \place {};
  
   %
   % Place normal peers
  \foreach \distxbetweengraphs\distybetweengraphs\graphid in {{0/3/0},{4/3/1},{2/0/2}}
  \foreach\place/\name  in {{(1+\distxbetweengraphs,-0.5+\distybetweengraphs)/ad\graphid}, {(1.5+\distxbetweengraphs,0+\distybetweengraphs)/bd\graphid}, {(0.6+\distxbetweengraphs,1.2+\distybetweengraphs)/cd\graphid}, {(2+\distxbetweengraphs,1+\distybetweengraphs)/dd\graphid},
                        {(2+\distxbetweengraphs,0+\distybetweengraphs)/ed\graphid}}
    \node[peers3] (\name) at \place {};

   \foreach \source/\dest in { ar0/cr0, bd0/cd0,  br0/cr0, cr0/dr0,ad0/ed0,dr0/er0,
   cd0/dd0,dr0/ad0}
    \path (\source) edge[edge1] (\dest);
    \foreach \source/\dest in {ar1/er1,bd1/dd1, cr1/dr1,bd1/ed1,
   cr1/dr1,dr1/bd1,ed1/ad1}
    \path (\source) edge[edge2] (\dest);
    \foreach \source/\dest in {ar2/er2,bd2/dd2,bd2/ed2, ad2/ed2,dr2/er2,cr2/dr2,
   cd2/dd2,dr2/ad2,br2/ar2}
    \path (\source) edge[edge3] (\dest);
    \foreach \source/\dest in {ar0/bd0,ad0/dd0,dd1/er1,cd2/ed2, cd1/ed1,br1/bd1}
    \path (\source) edge[edge4] (\dest);
    %%%%
    \node[label={[font=\footnotesize]0:\textbf{Neighborhood aggregation module}}] at (0.6,5) (title) {};
  %\node at (-0.8,3.6) (nA2) {};
  %  \node[label={[font=\footnotesize]0: : $K=2$-hop neighborhood}] at (0,3.6) (nB2) {};
  %  \path (nA2) edge[edge4] (nB2);
    %%%%%%%%
    \node at (0,2.35) (s11) {};
    \node at (-0.5,1.6) (s21) {\footnotesize{$\diffusedfeatvec\nodeshiftnot{\nodeind}{1}\layernot{0}$}};
    \draw[->,my line style4,to path={-| (\tikztotarget)}]
        (s11) edge (s21);
     \node at (6,2.7) (s12) {};
    \node at (5.8,1.8) (s22) {\footnotesize{$\diffusedfeatvec\nodeshiftnot{\nodeind}{2}\layernot{0}$}};
    \draw[->,my line style4,to path={-| (\tikztotarget)}]
        (s12) edge (s22);
    \node at (1.5,-0.4) (s13) {};
    \node at (0.8,0.2) (s23) {\footnotesize{\color{black}$\diffusedfeatvec\nodeshiftnot{\nodeind}{3}\layernot{0}$}};
    \draw[->,my line style4,to path={-| (\tikztotarget)}]
        (s13) edge (s23);
   %%%%%%%
   %\draw[my line style4] \convexpath{er0,dr0,cr0,br0,ar0}{8pt};
   %\draw[my line style4] \convexpath{br1,dr1,dd1,ed1,bd1}{8pt};
   %\draw[my line style4] \convexpath{er2,br2,ar2}{8pt};
   \newcommand{\myfontnow}{\tiny}
   \node at (0.5,3.1) (nA) {\myfontnow $\featvec_\nodeind$};
   \node at (4.5,3.1) (nB) {\myfontnow $\featvec_\nodeind$};
   \node at (2.5,0.1) (nC) {\myfontnow $\featvec_\nodeind$};
      \node at (1,2.5) (nA) {\myfontnow $\featvec_{\nodeind_{\scalebox{.8} 9}}$};
   \node at (5,2.5) (nB) {\myfontnow $\featvec_{\nodeind_{\scalebox{.8}9}}$};
   \node at (3,-0.5) (nC) {\myfontnow $\featvec_{\nodeind_{\scalebox{.8}9}}$};
   \node at (0,2.5) (nA) {\myfontnow $\featvec_{\nodeind_{\scalebox{.8}1}}$};
   \node at (4,2.5) (nB) {\myfontnow $\featvec_{\nodeind_{\scalebox{.8}1}}$};
   \node at (2,-0.5) (nC) {\myfontnow $\featvec_{\nodeind_{\scalebox{.8}1}}$};
    \node at (1.5,3) (nA) {\myfontnow $\featvec_{\nodeind_{\scalebox{.8}2}}$};
   \node at (5.5,3) (nB) {\myfontnow $\featvec_{\nodeind_{\scalebox{.8}2}}$};
   \node at (3.5,0) (nC) {\myfontnow $\featvec_{\nodeind_{\scalebox{.8}2}}$};
    \node at (0.6,4.2) (nA) {\myfontnow $\featvec_{\nodeind_{\scalebox{.8}3}}$};
   \node at (4.6,4.2) (nB) {\myfontnow $\featvec_{\nodeind_{\scalebox{.8}3}}$};
   \node at (2.6,1.2) (nC) {\myfontnow $\featvec_{\nodeind_{\scalebox{.8}3}}$};
    \node at (2,4) (nA) {\myfontnow $\featvec_{\nodeind_{\scalebox{.8}4}}$};
   \node at (6,4) (nB) {\myfontnow $\featvec_{\nodeind_{\scalebox{.8}4}}$};
   \node at (4,1) (nC) {\myfontnow $\featvec_{\nodeind_{\scalebox{.8}4}}$};
    \node at (2,3) (nA) {\myfontnow $\featvec_{\nodeind_{\scalebox{.8}5}}$};
   \node at (6,3) (nB) {\myfontnow $\featvec_{\nodeind_{\scalebox{.8}5}}$};
   \node at (4,0) (nC) {\myfontnow $\featvec_{\nodeind_{\scalebox{.8}5}}$};
    \node at (0,4) (nA) {\myfontnow $\featvec_{\nodeind_{\scalebox{.8}6}}$};
   \node at (4,4) (nB) {\myfontnow $\featvec_{\nodeind_{\scalebox{.8}6}}$};
   \node at (2,1) (nC) {\myfontnow $\featvec_{\nodeind_{\scalebox{.8}6}}$};
    \node at (-0.5,3.6) (nA) {\myfontnow $\featvec_{\nodeind_{\scalebox{.8}7}}$};
   \node at (3.5,3.6) (nB) {\myfontnow $\featvec_{\nodeind_{\scalebox{.8}7}}$};
   \node at (1.5,0.6) (nC) {\myfontnow $\featvec_{\nodeind_{\scalebox{.8}7}}$};
    \node at (-0.5,3) (nA) {\myfontnow $\featvec_{\nodeind_{\scalebox{.8}8}}$};
   \node at (3.5,3) (nB) {\myfontnow $\featvec_{\nodeind_{\scalebox{.8}8}}$};
   \node at (1.5,0) (nC) {\myfontnow $\featvec_{\nodeind_{\scalebox{.8}8}}$};
  
  % \node at (-0.5,-2) (J){};
  
  %\node at (6.5,6) (L){};
  \node at (1,4) (J){};
  \node at (4.8,6) (L){};
  %\draw[red,thick,dotted] 
\draw [fill=orange,fill opacity=0.1,draw=none,rounded corners=0.1cm]($(J.north west)+(-0.3,0.6)$) rectangle ($(L.south east)+(0.3,-0.6)$);
  %\draw[red,thick,dotted] 
 %\draw [fill=orange,fill opacity=0.1,draw=none,rounded corners=0.5cm]($(J.north west)+(-0.3,0.6)$) rectangle ($(L.south east)+(0.3,-0.6)$);
    %\draw[thick,dotted]     ($(I.north west)+(-0.5,0.15)$) rectangle ($(L.south east)+(0.5,-0.15)$);

\end{tikzpicture}
    \caption{NAM combines features using the multi-relational graph. With reference to node $n$, and aggregation of $R$-hop neighbors (here $R=2$), note that the local neighborhood is not the same across different graphs.}
    \label{fig:neighboragrmodule}
\end{figure}

\noindent\textbf{Neighborhood aggregation module (NAM)}.
Consider a neighborhood aggregation module per relation and per node, that combines linearly the information available locally in each neighborhood. Since the neighborhood depends on the particular relation $i$ and node $n$, we have (cf. \eqref{eq:neighborhood}) 
\begin{align}
\mathbf{h}_{{n}{i}}^{(l)}
:=
\sum_{{n'}\in\mathcal{N}_{n}^{({i})}} A_{nn'i}
\check{\mathbf{z}}_{{n'}{i}}^{(l-1)}.
\label{eq:sem}
\end{align}
While the entries of 
$\mathbf{h}_{{n}{i}}^{(l)}$ depend 
only on the one-hop neighbors of $n$ (one-hop diffusion), successive application of this operation across layers will expand the diffusion reach, eventually spreading the information across the network. Letting $\mathbf{A}_{i}^{(r)}:=\mathbf{A}_{i}^r$ denote the $r$th power of feature matrices for $r=1,\ldots,R$ and ${i}=1,\ldots,{I}$, vectors $\mathbf{A}_i^r\mathbf{x}$ hold linear combinations of the values of $\mathbf{x}$ in the $r$-hop neighborhood~\cite{shuman2013emerging}; thus, \eqref{eq:sem} becomes
\begin{align}
\mathbf{h}_{{n}{i}}^{(l)}
=\sum_{{r=1^{\hspace{.015cm} }}}^R
\sum_{n'=1}^{N} C^{(r,l)}_{i}A_{nn'i}^{(r)}
\check{\mathbf{z}}_{{n'}{i}}^{(l-1)}
\label{eq:gf}
\end{align}
where the learnable coefficients $C^{(r,l)}_{i}$ weigh the corresponding $r$th hop neighbors of node $n$ according to relation ${i}$. Per layer $l$,  $\{C^{(r,l)}_{i}\}_{\forall (r,i)}$ are collected in the $R\times{I}$ matrix
$\mathbf{C}^{(l)}$. The proposed transformation in \eqref{eq:gf} aggregates the diffused signal in the $R$-hop neighborhoods per $i$; see also Fig. \ref{fig:neighboragrmodule}. 

%vector An alternative to account for multiple hops is to apply successively \eqref{eq:sem} within one layer, which is left as future work.

\noindent\textbf{Graph adaptive module (GAM)}.
Feature vector $\mathbf{h}_{{n}{i}}^{(l)}$ captures the diffused input per relation ${i}$, and its role will depend on the inference task at hand. In predicting voting {preference,} for instance, the friendship network may be more important than the coworker relation; cf. Fig.~\ref{fig:multilayer}. As a result, the learning algorithm should be able to adapt to the prevalent features. This motivates the weighted combination 
%of $\mathbf{h}_{{n}{i}}^{(l)}$ across $i$ to obtain 
%Next, to mix features in the graph data, we combine the entries in \todo{change motivation}
%$\mathbf{h}_{{n}{i}}^{(l)}$ using (trainable) parameters as follows % Specifically, we propose to compute 
\begin{align}\label{eq:graphadaptive}
\graphadaptivec_{{n}{i}}^{(l)}:=&\sum_{{i'}=1}^{I}{R}_{{i}{i'}{n}}^{(l)}{\mathbf{h}_{{n}{i'}}^{(l)}}
\end{align}
where  
$\{R_{{i}{i'}{n}}^{(l)}\}$ mix features of graphs $i$ and $i'$. Collecting weights 
$\{{R}_{{i}{i'}{n}}^{(l)}\}$ ${\forall ({i},{i'},{n})}$, yields the trainable ${I}\times{I}\times N$ tensor $\underline{\mathbf{R}}^{(l)}$. The graph-mixing weights enable our TGCN to learn how to combine and adapt across different relations encoded by the multi-relational graph; see also Fig.~\ref{fig:graphagrmodule}. 

Clearly, if prior information on the dependence among relations is {available}, this can be used to constrain the structure $\underline{\mathbf{R}}^{(l)}$ to be e.g., diagonal or sparse.  The graph-adaptive combination in \eqref{eq:graphadaptive} allows for different $ R_{ii'n}$ per $n$. Considering the same $ R$ for each $n$, that is $ R_{ii'n}^{(l)}= R_{ii'}^{(l)}$, results in a design with {less} parameters at the expense of reduced flexibility. For example, certain voters may be affected more {by} their friends, whereas others {by} their coworkers. Using the GAM, our network can achieve personalized predictions.
\begin{figure}
    \centering
    % Layer 1 = 17,23,20
% Layer 2 = 9, 9, 42

\begin{tikzpicture}[ transform shape]
\definecolor{paramcolor}{HTML}{50ae55}
\definecolor{inputcolor}{HTML}{fd9727}
\definecolor{outputcolor}{HTML}{f1453d}

% Height scale = 7 = 0.1

\tikzset{
	inputmat/.style = {rectangle, draw=inputcolor!70, fill=inputcolor!40, thick, minimum width=1.4cm, minimum height = 0.7cm},
	output1mat1/.style = {rectangle, draw=purple, fill=purple, line width=0.1mm, inner sep=0pt, minimum width=0.15cm, minimum height = 1.4cm},
	output2mat1/.style = {rectangle, draw=blue, fill=blue, line width=0.1mm, inner sep=0pt, minimum width=0.15cm, minimum height = 1.4cm},
	output3mat1/.style = {rectangle, draw=black, fill=black, line width=0.1mm, inner sep=0pt, minimum width=0.15cm, minimum height = 1.4cm},
	output1mat2/.style = {rectangle, draw=outputcolor!90, fill=outputcolor!70,  line width=0.1mm, inner sep=0pt, minimum width=1.0cm, minimum height = 0.0429cm},
	output2mat2/.style = {rectangle, draw=outputcolor!90, fill=outputcolor!30, line width=0.1mm, inner sep=0pt, minimum width=1.0cm, minimum height = 0.0858cm},
	output3mat2/.style = {rectangle, draw=outputcolor!90, fill=outputcolor!50, line width=0.1mm, inner sep=0pt, minimum width=1cm, minimum height = 0.0429cm},
	thickermat/.style = {rectangle, draw=black!50, fill=black!20, thick, minimum width=1.4cm, minimum height = 1cm},
	thinmat/.style = {rectangle, draw=black!50, fill=black!20, thick, minimum width=0.2cm,
		minimum height = 1cm},
	parameter_1_1/.style = {rectangle, draw=paramcolor!70, fill=paramcolor!40, thin, minimum width=0.7cm, minimum height = 0.243cm}, % 17
	parameter_1_2/.style = {rectangle, draw=paramcolor!70, fill=paramcolor!40, thin, minimum width=0.7cm, minimum height = 0.329cm}, % 23
	parameter_1_3/.style = {rectangle, draw=paramcolor!70, fill=paramcolor!40, thin, minimum width=0.7cm, minimum height =0.286cm}, % 20
	parameter_2_1/.style = {rectangle, inner sep=0pt, draw=paramcolor!70, fill=paramcolor!40, thin, minimum width=0.7cm, minimum height = 0.0429cm},   % 3
	parameter_2_2/.style = {rectangle, inner sep=0pt, draw=paramcolor!70, fill=paramcolor!40, thin, minimum width=0.7cm, minimum height = 0.0858cm},   % 6
	parameter_2_3/.style = {rectangle,
	inner sep=0pt,
	draw=paramcolor!70, fill=paramcolor!40, thin, minimum width=0.7cm, minimum height = 0.0429cm},   % 3   
	parameter/.style = {rectangle, draw=paramcolor!70, fill=paramcolor!40, thick, minimum width=0.4cm, minimum height = 0.4cm},
	output/.style = {rectangle, fill=blue,fill opacity=0.4,draw=none , thick, minimum width=0.15cm, minimum height = 1.4cm},
}
\node [output1mat1] (yy1) at (0.9,2)  {};
\node [output2mat1] (yy2) at (1.3,2) {};
\node [output3mat1] (yy3) at (1.7,2) {};
%\draw (wmult2) edge[out=0,in=180,->] (yy2);

\node [output] (y) at (5,2) {};

\node [output3mat1,minimum size=0.25cm] (R1) at (3,1.6) {};
\node [output2mat1,minimum size=0.25cm] (R2) at (3,2) {};
\node [output1mat1,minimum size=0.25cm] (R3) at (3,2.4) {};
\node [] (Rlab) at (3.5,1.6) {{\tiny${R}_{i3n}^{(l)}$}};
\node [] (Rlab) at (3.5,2) {{\tiny${R}_{i2n}^{(l)}$}};
\node [] (Rlab) at (3.5,2.4) {{\tiny${R}_{i1n}^{(l)}$}};
\node []() at (0.8,3)
[]{\footnotesize{$\diffusedfeatvec\nodeshiftnot{\nodeind}{1}\layernot{l}$}}; % top label
\node []() at (1.3,3)
[]{\footnotesize{$\diffusedfeatvec\nodeshiftnot{\nodeind}{2}\layernot{l}$}}; % top label
\node []() at (1.8,3)
[]{\footnotesize{$\diffusedfeatvec\nodeshiftnot{\nodeind}{3}\layernot{l}$}}; % top label

\node []() at (5,3)
[]{\footnotesize{\color{black}$\graphadaptivec\nodeshiftnot{\nodeind}{\shiftind}\layernot{\layerind}$}};

\node [](mult) at (2.4,2) {$\mathbf{\times}$};

\node [](eq) at (4.3,2) {$\mathbf{=}$};

%\draw[-{>}] (yy1) -- (mult);
%\draw[-{>}] (R) -- (mult);
%\draw[-{>}] (yy3) -- (y);
\node[label={[font=\footnotesize]0:\textbf{ Graph adaptive module}}] at (1.2,4) (title) {};

\node at (0.8,2.9) (mJ1){};
\node at (0.8,1.3) (mJ){};
  \node at (3.5,1.3) (mL1){};
  \node at (3.5,2.9) (mL){};
%\draw \convexpath{mJ,mJ1,mL,mL1}{10pt};

 \node at (1.8,3) (J){};
  \node at (4.6,5) (L){};
  %\draw[red,thick,dotted] 
\draw [fill=blue,fill opacity=0.1,draw=none,rounded corners=0.1cm]($(J.north west)+(-0.3,0.6)$) rectangle ($(L.south east)+(0.3,-0.6)$);

\end{tikzpicture}
    \caption{GAM combines the features per $i$, based on the trainable coefficients $\{{R}_{{i}{i'}{n}}\}$. When $\underline{\mathbf{R}}$ is sparse only features corresponding to the most significant relations will be active.}
    \label{fig:graphagrmodule}
\end{figure}

\noindent\textbf{Feature aggregation module (FAM)}.
Next, the extracted GAM features are mixed using learnable scalars
$W_{nipp'} ^{(l)}$ as 
\begin{align}\label{eq:linconv}
\underline{{Z}}_{{n}{i}p}^{(l)}:=&
\sum_{p'=1}^{P^{(l-1)}}W_{nipp'}^{(l)}G_{nip'}^{(l)}
\end{align}
for all $(n,i,p)$, where $G_{nip'}^{(l)}$ represents the $p'$th entry of $\graphadaptivec_{{n}{i}}^{(l)}$. 
%and $W_{nip'}^{(l)}$ are learnable parameters mixing the features of the $l$th layer.
The ${N}\times{I}\times P^{(l)}\times P^{(l-1)}$ tensor $\weighttensor^{(l)}$ collects the feature mixing weights 
$\{W_{nipp'}^{(l)}\}_{\forall (n,i,p,p')}$.

The linear modules that map the input tensor $\check{\underline{\mathbf{Z}}}^{(l-1)}$ to $\underline{\mathbf{Z}}^{(l)}$ can be now summarized as follows 
\begin{align}
\label{eq:linGCN}
\underline{\mathbf{Z}}^{(l)}&:=
f(\check{\underline{\mathbf{Z}}}^{(l-1)};
\bm{\theta}_z^{(l)}),\;\;\;\text{with}\\
\label{eq:param}
\bm{\theta}_z^{(l)}&:=[\text{vec}(\weighttensor^{(l)});\text{vec}(\underline{\mathbf{R}}^{(l)})
;\text{vec}(\mathbf{C}^{(l)})]^{\top}
\end{align}
where $f$ denotes the synthesis of the three linear modules introduced (namely NAM, GAM and FAM), while $\bm{\theta}_z^{(l)}$ collects the learnable weights involved in those modules [cf. \eqref{eq:gf}-\eqref{eq:linconv}].

\subsection{Residual GCN layer}
Successive application of $L$ TGCN layers diffuses the input ${\mathbf{X}}$ 
across the $LR$-hop graph neighborhood, cf.~\eqref{eq:sem}. However, the exact 
size of the relevant neighborhood is not always known a priori. To endow our architecture with increased flexibility, we propose a residual TGCN layer that inputs ${\mathbf{X}}$ at each $l$, and thus can include ``sufficient data statistics'' that may have been lost after successive  diffusions. This `raw data reuse' is also known as a skip connection~\cite{he2016deep,ruiz2019gated}. Skip connections also emerge when an optimization solver is `unrolled' as a deep neural network with each layer having the form of an iteration; see also~\cite{zhang2019real}. Specifically, the linear operation in \eqref{eq:linGCN} is replaced by the residual linear tensor mapping \cite[Ch. 10]{goodfellow2016deep}
\begin{align}
\underline{\mathbf{Z}}^{(l)}:=
f(\check{\underline{\mathbf{Z}}}^{(l-1)};
\bm{\theta}_z^{(l)})+
f({\mathbf{X}};
\bm{\theta}_x^{(l)})
\label{eq:residuallayer}
\end{align}
where $\bm{\theta}_x^{(l)}$ collects trainable parameters as those in \eqref{eq:param}. 
When viewed as a transformation from  ${\mathbf{X}}$ to $\underline{\mathbf{Z}}^{(l)}$, the operator in  \eqref{eq:residuallayer} implements a broader class of graph diffusions than the one in \eqref{eq:linGCN}. If, for example, $l=3$ and $k=1$, then the first summand in \eqref{eq:residuallayer} is a 1-hop diffusion of a signal that corresponded to a $2$-hop (nonlinear) diffused version of ${\mathbf{X}}$, while the second summand diffuses ${\mathbf{X}}$ in one hop. At a more intuitive level, the presence of the second summand also guarantees that the impact of ${\mathbf{X}}$ in the output does not vanish as the number of layers grow. The autoregressive mapping in \eqref{eq:residuallayer} facilitates {the} application of our architecture with time-varying inputs and labels. Specifically, with $t$ indexing time and given time-varying data $\{{\mathbf{X}}_t\}_{t}^T$, one would set $l=t$, replace ${\mathbf{X}}$ in \eqref{eq:residuallayer} with ${\mathbf{X}}^{(l)}$, and set ${\mathbf{X}}^{(l)}={\mathbf{X}}_t$. This will be studied in detail in our future 
work towards predicting dynamic processes over multi-relational graphs. 
\subsection{Initial and final layers}
 Regarding layer $l=1$, its input $\check{\underline{\mathbf{Z}}}^{(0)}$ is 
\begin{align}\label{eq:input_first_layer}
\check{\mathbf{z}}_{{n}{i}} ^{(0)}=\mathbf{x}_{n} \;\;\text{for}\;\; \text{all}\;\; (n,i).
\end{align}
AT the other end, the output of our graph architecture is obtained by taking the output of layer $l=L$, and applying 
\begin{align}\label{eq:output}
        \hat{\mathbf{Y}}:=g(\check{\underline{\mathbf{Z}}}
        ^{(L)};\bm{\theta}_g)
        \end{align}
where $g(\cdot)$ is a nonlinear function, $\hat{\mathbf{Y}}$ is an ${N}\times K$ matrix, 
$\hat{Y}_{{n},k}$ represents the probability that $y_{n}=k$, and $\bm{\theta}_g$ collects trainable parameters. Function $g(\cdot)$ depends on the specific application, with the normalized exponential function (softmax) being a popular choice for classification problems; that is,
\begin{align}
 \hat{Y}_{{n},k}=\frac{\exp{\check{\underline{{Z}}}_{n,k}^{(L)}}}{
 \sum_{k=1}^K\exp{\check{\underline{{Z}}}_{n,k}^{(L)}}}.
\end{align}

\begin{figure*}
    \centering
    \begin{tikzpicture}[%
 block/.style={rectangle, fill=gray, text centered, minimum height=8mm, rounded corners=1mm}
        ,my line style1/.style={ draw=purple,thick}
        ,my line style2/.style={ draw=blue,thick}
        ,my line style3/.style={ draw=black,thick}
        ,every label/.append style = {label distance=1pt, inner sep=1pt, 
                             align=left, font=\tiny}
        %,every edge/.style={my line style}
        ,my edge/.style={my line style, mylabel=#1}
        ,my curved edge/.style n args={4}{my line style, out=#1, in=#2, looseness=#3, mylabel=#4}
        ,edge1/.style={my line style1,dashed}
        ,mylabel1/.style={my label style{#1}}
        ,edge2/.style={my line style2, densely dotted}
        ,edge3/.style={my line style3}
        ,mylabel/.style={edge node={node[my label style]{#1}}}]
    %%%%%%%%%%%%%%
   %NN input
   %%%%%%%%%%%%%%
     % Place super peers and connect them
     
\foreach \distxbetweengraphs\distybetweengraphs\graphid in {{0/-1/0},{0/0/1},{0/1/2}}
  \foreach \place/\name in {{(0+\distxbetweengraphs,0+\distybetweengraphs)/ar\graphid}, {(0.2+\distxbetweengraphs,0.1+\distybetweengraphs)/br\graphid}, {(0+\distxbetweengraphs,0.5+\distybetweengraphs)/cr\graphid}, {(-0.2+\distxbetweengraphs,0.4+\distybetweengraphs)/dr\graphid},
           {(-0.2+\distxbetweengraphs,0+\distybetweengraphs)/er\graphid}}
    \node[superpeers2] (\name) at \place {};
  
   %
   % Place normal peers
  \foreach \distxbetweengraphs\distybetweengraphs\graphid in {{0/-1/0},{0/0/1},{0/1/2}}
  \foreach\place/\name  in {{(0.5+\distxbetweengraphs,-0.2+\distybetweengraphs)/ad\graphid}, {(0.8+\distxbetweengraphs,0+\distybetweengraphs)/bd\graphid}, {(0.3+\distxbetweengraphs,0.3+\distybetweengraphs)/cd\graphid}, {(0.8+\distxbetweengraphs,0.3+\distybetweengraphs)/dd\graphid},
                        {(0.8+\distxbetweengraphs,0+\distybetweengraphs)/ed\graphid}}
    \node[peers2] (\name) at \place {};

   \foreach \source/\dest in {ar0/bd0, ar0/cr0, ad0/dd0,bd0/cd0,  br0/cr0, cr0/dr0,ad0/ed0,dr0/er0,
   cd0/dd0,dr0/ad0}
    \path (\source) edge[edge1] (\dest);
    \foreach \source/\dest in {ar1/er1,bd1/dd1,cd1/ed1, cr1/dr1,bd1/ed1,dd1/er1,
   cr1/dr1,dr1/bd1,br1/bd1,ed1/ad1}
    \path (\source) edge[edge2] (\dest);
    \foreach \source/\dest in {ar2/er2,bd2/dd2,cd2/ed2, cr2/dr2,bd2/ed2, cr2/dr2,ad2/ed2,dr2/er2,
   cd2/dd2,dr2/ad2,br2/ar2}
    \path (\source) edge[edge3] (\dest);
   %%%%%%%%%%%%%%
   %NN architecture
   %%%%%%%%%%%%%%
    \node[nnblock] at (3,0) (nA) {};
    \node[nam,label={[label distance=-0.3cm,text depth=-2ex,rotate=-90]right:\tiny{\textbf{NAM}}}] at (2.7,0) (namA){};
    \node[gam,label={[label distance=-0.3cm,text depth=-2ex,rotate=-90]right:\tiny{\textbf{GAM}}}] at (3,0) (gamA){};
    \node[fam,label={[label distance=-0.3cm,text depth=-2ex,rotate=-90]right:\tiny{\textbf{FAM}}}] at (3.3,0) (famA){};
    \node[nnblock] at (3+\distbetweenlayers,0) (nB) {};
        \node[nam,label={[label distance=-0.3cm,text depth=-2ex,rotate=-90]right:\tiny{\textbf{NAM}}}] at (2.7+\distbetweenlayers,0) (namA){};
    \node[gam,label={[label distance=-0.3cm,text depth=-2ex,rotate=-90]right:\tiny{\textbf{GAM}}}] at (3+\distbetweenlayers,0) (gamA){};
    \node[fam,label={[label distance=-0.3cm,text depth=-2ex,rotate=-90]right:\tiny{\textbf{FAM}}}] at (3.3+\distbetweenlayers,0) (famA){};
    \node[nnblock] at (3+2*\distbetweenlayers,0) (nC) {};
        \node[nam,label={[label distance=-0.3cm,text depth=-2ex,rotate=-90]right:\tiny{\textbf{NAM}}}] at (2.7+2*\distbetweenlayers,0) (namA){};
    \node[gam,label={[label distance=-0.3cm,text depth=-2ex,rotate=-90]right:\tiny{\textbf{GAM}}}] at (3+2*\distbetweenlayers,0) (gamA){};
    \node[fam,label={[label distance=-0.3cm,text depth=-2ex,rotate=-90]right:\tiny{\textbf{FAM}}}] at (3.3+2*\distbetweenlayers,0) (famA){};
    \node[] at (3+3*\distbetweenlayers,0) (ndots) {\huge{$\cdots$}};
    \node[nnblock] at (3+4*\distbetweenlayers,0) (nD) {};
        \node[nam,label={[label distance=-0.3cm,text depth=-2ex,rotate=-90]right:\tiny{\textbf{NAM}}}] at (2.7+4*\distbetweenlayers,0) (namA){};
    \node[gam,label={[label distance=-0.3cm,text depth=-2ex,rotate=-90]right:\tiny{\textbf{GAM}}}] at (3+4*\distbetweenlayers,0) (gamA){};
    \node[fam,label={[label distance=-0.3cm,text depth=-2ex,rotate=-90]right:\tiny{\textbf{FAM}}}] at (3.3+4*\distbetweenlayers,0) (famA){};
    \node[outblock,label={[label distance=-0.6cm,text depth=-2ex,rotate=-90]right:\tiny{\textbf{Output layer}}}] at (3+5*\distbetweenlayers,0) (nE) {};
    \definecolor{paramcolor}{HTML}{50ae55}
\definecolor{inputcolor}{HTML}{fd9727}
\definecolor{outputcolor}{HTML}{f1453d}
\tikzset{
	inputmat/.style = {rectangle, draw=inputcolor!70, fill=inputcolor!40, thick, minimum width=1.4cm, minimum height = 1cm},
	outputmat/.style = {rectangle, draw=outputcolor!0, fill=outputcolor!50, thick, minimum width=0.6cm, minimum height = 1.2cm},
	thickermat/.style = {rectangle, draw=black!50, fill=black!20, thick, minimum width=1.4cm, minimum height = 1cm},
	parameter/.style = {rectangle, draw=paramcolor!70, fill=paramcolor!40, thick, minimum width=0.6cm, minimum height = 1.4cm},
}
    \node [outputmat,label={[label distance=-0.5cm,text depth=0ex]right:\footnotesize$\predictionmat$}] at (3+5*\distbetweenlayers+1,0) (nF) {};
    
    \draw [line width=0.1mm] ($(nF.north west) + (0,0.1) $ ) -- ($(nF.north east) + (0,0.1)$); % horizontal line
    \draw [line width=0.1mm] ($(nF.north west) + (0,0.07) $ ) -- ($(nF.north west) + (0,0.13)$); % left tick
\draw [line width=0.1mm] ($(nF.north east) + (0,0.07) $ ) -- ($(nF.north east) + (0,0.13)$); % bottom tick
\node [above=-0.05cm of nF]()[]{\footnotesize $^\nbrclasses$}; % top label

\draw [line width=0.1mm] ($(nF.north east) + (0.1,0) $ ) -- ($(nF.south east) + (0.1,0.0)$); % vertical line
\draw [line width=0.1mm] ($(nF.north east) + (0.13,0) $ ) -- ($(nF.north east) + (0.07,0)$); % top tick
\draw [line width=0.1mm] ($(nF.south east) + (0.13,0) $ ) -- ($(nF.south east) + (0.07,0)$); % bottom tick
\node [right=0cm of nF]()[]{\footnotesize $^{\nbrnodes}$}; % left label

    \newcommand{\xposrel}{-1cm}
    \newcommand{\yposrel}{0.3cm}
    \node[above right = \yposrel and  \xposrel of nA] (eA) {\footnotesize $l=1$};
    \node[above right = \yposrel and  \xposrel of nB] (eB) {\footnotesize $l=2$};
    \node[above right = \yposrel and  \xposrel of nC] (eC) {\footnotesize $l=3$};
    \node[above right = \yposrel and  \xposrel of nD] (eD) {\footnotesize $l=\nbrlayers$};

    \node at (0.3,-2.5) (inp){$\datamatrix$};
    
    \draw[->] (nA.east) -- node [above] {\footnotesize $\outputlaytensor\layernot{1}$} (nB);
    \draw[->] (nB.east) -- node [above] {\footnotesize$\outputlaytensor\layernot{2}$} (nC);
    \draw[->] (nC.east) -- node [above] {\footnotesize$\outputlaytensor\layernot{3}$} (ndots);
    \draw[->] (ndots.east) --  (nD);
    \draw[->] (nD.east) -- node [above] {\footnotesize$\outputlaytensor\layernot{\nbrlayers}$}(nE);
    \draw[->] (nE.east) -- node [above] {}(nF);
    \draw[->, to path={-| (\tikztotarget)}]
        (inp) edge% bend the arrows [bend left]
        (nA) (inp) edge (nB) 
        (inp) edge (nC)(inp) edge (nD);

\end{tikzpicture}
    \caption{TGCN with $L$ hidden (black) and one output (red) layers. The input $\mathbf{X}$ contains a collection of features per node and the output to be predicted is the probability of each node {belonging} to each the $K$ classes (labels) considered. Each layer of the TGCN is composed of our three novel modules (NAM, GAM, FAM) described in equations \eqref{eq:gf}, \eqref{eq:graphadaptive}, and \eqref{eq:linconv}.  Notice the skip connections that input $\mathbf{X}$ to each layer [cf. \eqref{eq:residuallayer}].
    %\red{context}
    }
    \label{fig:grnn}
\end{figure*}
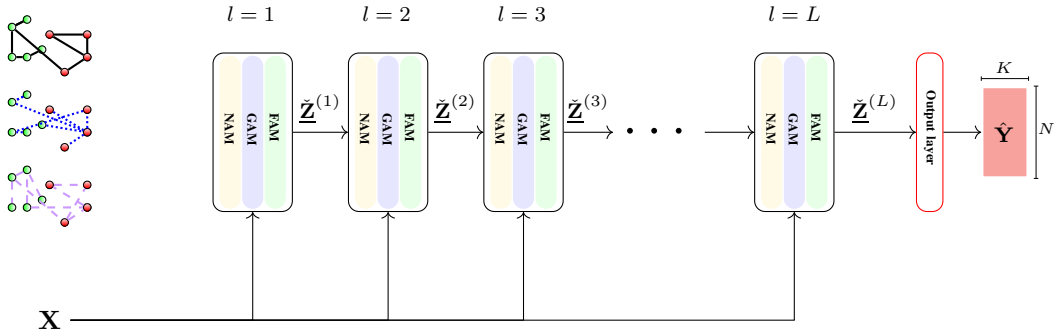

For notational convenience, the global mapping $\mathcal{F}$ from $\mathbf{X}$ to 
$\hat{\mathbf{Y}}$ dictated by our TGCN architecture is 
\begin{align}
 \hat{\mathbf{Y}}:=\mathcal{F}\big(\mathbf{X};\{\bm{\theta}_z^{(l)}\}_{l=1}^L,\{\bm{\theta}_x^{(l)}\}_{l=1}^L,\bm{\theta}_g\big)
\end{align} 
and it is summarized by the block diagram of Fig.~\ref{fig:grnn}.

\subsection{Training and graph-smooth regularizers}
The proposed architecture is parameterized by the weights in \eqref{eq:residuallayer} and \eqref{eq:output}. We learn these weights during the training phase by minimizing 
the discrepancy between the estimated and the given labels; that is, we solve 
\begin{align}\label{eq:trainobj}    \min_{\{\bm{\theta}_z^{(l)}\}_{l=1}^L,\{\bm{\theta}_x^{(l)}\}_{l=1}^L,\bm{\theta}_g}&
\mathcal{L}_{tr} (\hat{\mathbf{Y}},\mathbf{Y}) +\mu_1\sum_{{i}=1}^{I}\text{Tr}(\hat{\mathbf{Y}}^{\top}\mathbf{A}_{i}\hat{\mathbf{Y}})
\nonumber\\
+&\mu_2\regfun\big(\{\bm{\theta}_z^{(l)}\}_{l=1}^L,\{\bm{\theta}_x^{(l)}\}_{l=1}^L\big)
+\lambda \sum_{l=1}^L\|\underline{\mathbf{R}}^{(l)}\|_1\nonumber\\
\text{s.t.}&~~
\hat{\mathbf{Y}}=\mathcal{F}\big(
\mathbf{X};\{\bm{\theta}_z^{(l)}\}_{l=1}^L,\{\bm{\theta}_x^{(l)}\}_{l=1}^L,\bm{\theta}_g\big). 
\end{align}
For SSL, a reasonable choice for the fitting cost is the cross-entropy loss over the labeled examples, i.e., $\mathcal{L}_{tr} (\hat{\mathbf{Y}},\mathbf{Y}):=-\sum_{{n}\in\mathcal{M}}\sum_{k=1}^K 
Y_{{n} k}\ln{\hat{Y}_{{n} k}}$. 
%The regularizers in \eqref{eq:trainobj} are used to avoid overfitting and promote application-specific prior knowledge.

The first regularization term in \eqref{eq:trainobj} promotes smooth label estimates over the graphs \cite{smola2003kernels}, while the second $\regfun(\cdot)$ is an $\ell_2$ norm over the TGCN parameters typically used to avoid overfitting~\cite{goodfellow2016deep}. Finally, the $\ell_1$ norm in the third regularizer encourages learning sparse mixing coefficients, and hence it promotes activating only a subset of relations per $ l$. %ED graphs with a large number of perturbed edges will result to a higher cost function in \eqref{eq:trainobj}. 
The learning algorithm will assign larger combining weights to topologies that are most appropriate for the given data. A backpropagation algorithm~\cite{rumelhart1986learning} is employed to minimize \eqref{eq:trainobj}. The computational complexity of evaluating \eqref{eq:residuallayer} scales linearly with the number of nonzero entries in $\underline{\mathbf{A}}$ (edges) [cf. \eqref{eq:sem}].

To recap, while most prior GCN works entail a single graph with one type of diffusion \cite{bronstein2017geometric,kipf2016semi}, this section has introduced a (residual) TGCN that: i) accounts for multiple graphs over the same set of nodes; ii) diffuses signals across each of the different graphs; iii) combines the signals of the different graphs using adaptive (learnable) coefficients; iv) implements a simple but versatile residual  tensor map \eqref{eq:residuallayer}; and v) includes several types of graph-based regularizers. 

\section{Tensor graphs for robust GCNs}

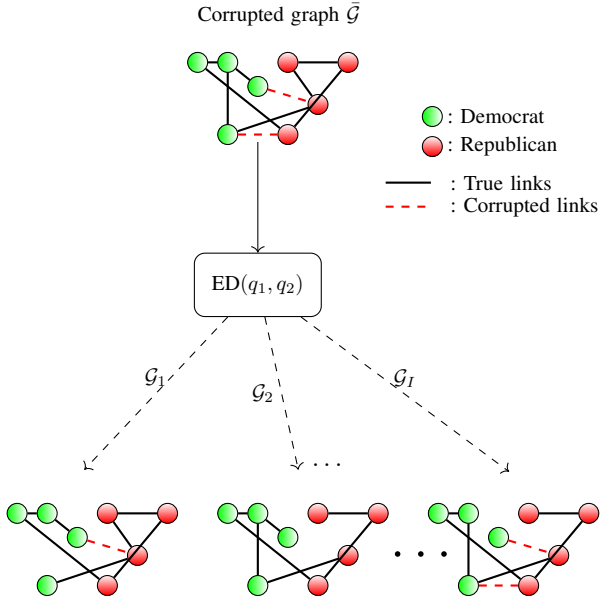
\begin{figure}
	\centering
	\begin{tikzpicture}[%
        scale=0.8, every node/.style={scale=0.8},
        block/.style={rectangle, fill=gray, text centered, minimum height=8mm, rounded corners=1mm}
        ,my line style1/.style={ draw=red,thick}
        ,my line style2/.style={ draw=blue,thick}
        ,my line style3/.style={ draw=black,thick}
        ,every label/.append style = {label distance=1pt, inner sep=1pt, 
                             align=left, font=\tiny}
        %,every edge/.style={my line style}
        ,my edge/.style={my line style, mylabel=#1}
        ,my curved edge/.style n args={4}{my line style, out=#1, in=#2, looseness=#3, mylabel=#4}
        ,edge1/.style={my line style1,dashed}
        ,mylabel1/.style={my label style{#1}}
        ,edge2/.style={my line style2, densely dotted}
        ,edge3/.style={my line style3}
        ,mylabel/.style={edge node={node[my label style]{#1}}}]
  % Place super peers and connect them
\newcommand{\disttopgraphx}{-2}
\newcommand{\disttopgraphy}{+3}
\foreach \place/\name in {{(5+\disttopgraphx,-0.5+\disttopgraphy)/ar0}, {(5.5
+\disttopgraphx,0.3+\disttopgraphy)/br0}, {(5+\disttopgraphx,0.7+\disttopgraphy)/cr0}, {(4.5+\disttopgraphx,0.7+\disttopgraphy)/dr0}}
    \node[superpeers1] (\name) at \place {};
\foreach\place/\name  in {{(6+\disttopgraphx,-0.5+\disttopgraphy)/ad0}, {(6.5+\disttopgraphx,0+\disttopgraphy)/bd0}, {(6+\disttopgraphx,0.7+\disttopgraphy)/cd0}, {(7+\disttopgraphx,0.7+\disttopgraphy)/dd0}}
\node[peers1] (\name) at \place {};

\foreach \distxbetweengraphs\distybetweengraphs\graphid in {{0/-4.5/1},{3.5/-4.5/2},{7/-4.5/3}}
  \foreach \place/\name in {{(0+\distxbetweengraphs,-0.5+\distybetweengraphs)/ar\graphid}, {(0.5+\distxbetweengraphs,0.3+\distybetweengraphs)/br\graphid}, {(0+\distxbetweengraphs,0.7+\distybetweengraphs)/cr\graphid}, {(-0.5+\distxbetweengraphs,0.7+\distybetweengraphs)/dr\graphid}}
    \node[superpeers1] (\name) at \place {};
  
   %
   % Place normal peers
  \foreach \distxbetweengraphs\distybetweengraphs\graphid in  {{0/-4.5/1},{3.5/-4.5/2},{7/-4.5/3}}
  \foreach\place/\name  in {{(1+\distxbetweengraphs,-0.5+\distybetweengraphs)/ad\graphid}, {(1.5+\distxbetweengraphs,0+\distybetweengraphs)/bd\graphid}, {(1+\distxbetweengraphs,0.7+\distybetweengraphs)/cd\graphid}, {(2+\distxbetweengraphs,0.7+\distybetweengraphs)/dd\graphid}}
    \node[peers1] (\name) at \place {};

   \foreach \source/\dest in {ar0/bd0, ar0/cr0, ad0/dd0,bd0/cd0,  br0/cr0, cr0/dr0,
   cd0/dd0,dr0/ad0}
    \path (\source) edge[edge3] (\dest);
    \path (ar0) edge[edge1] (ad0);
    \path (br0) edge[edge1] (bd0);

     \foreach \source/\dest in {ar1/bd1, ad1/dd1,bd1/cd1,  br1/cr1, cr1/dr1,
   cd1/dd1,dr1/ad1}
    \path (\source) edge[edge3] (\dest);
    \path (br1) edge[edge1] (bd1);

  \foreach \source/\dest in {ar2/bd2, ar2/cr2, cr2/dr2,ad2/dd2,br2/cr2,
   cd2/dd2,dr2/ad2}
    \path (\source) edge[edge3] (\dest);
    %\path (ar2) edge[edge1] (ad2);
    
    \foreach \source/\dest in {ar3/bd3, ar3/cr3, ad3/dd3, cr3/dr3,
   cd3/dd3,dr3/ad3}
    \path (\source) edge[edge3] (\dest);
    \path (ar3) edge[edge1] (ad3);
    \path (br3) edge[edge1] (bd3);
    \node[esl] at (3.5,0) (esl1) {\normalsize ED$(q_1,q_2)$};
  %labels 
  \node[label={[font=\normalsize]0: Corrupted graph $\bar{\mathcal{G}}$}] at (4.3+\disttopgraphx,1.5+\disttopgraphy) (cg) {};
   \node[] at (5.5+\disttopgraphx,-0.5+\disttopgraphy) (cg0) {};
  \node[] at (0.5,-3.2) (cg1) {};
  \node[] at (4.2,-3.2) (cg2) {};
  \node[]at (7.8,-3.2) (cg4) {};
  \node[label={[font=\normalsize]0: \large{$\cdots$}}] at (4.2,-3) (cg6) {};
  \node[label={[font=\normalsize]0: \Huge{$\cdots$}}] at (5.5,-4.5) (cg3) {};
  \draw[->] (cg0) --  (esl1);
  \draw[dashed,->] (esl1) -- node [above] {$\mathcal{G}_1$} (cg1);
   \draw[dashed,->] (esl1) -- node [left] {$\mathcal{G}_2$} (cg2);
    \draw[dashed,->] (esl1) -- node [above] {$\mathcal{G}_I$} (cg4);
  \node at (7.5+\disttopgraphx,1.7) (nA1) {};
    \node[label={[font=\normalsize]0: : True links}] at (8.5+\disttopgraphx,1.7) (nB1) {};
    \path (nA1) edge[edge3] (nB1);
  \node at (7.5+\disttopgraphx,1.3) (nA2) {};
    \node[label={[font=\normalsize]0: : Corrupted links}] at (8.5+\disttopgraphx,1.3) (nB2) {};
    \path (nA2) edge[edge1] (nB2);
     \node[superpeers,label={[font=\normalsize]0: : Democrat}] (d) at (8.4+\disttopgraphx,2.8) {};
    \node[peers,label={[font=\normalsize]0: : Republican}] (r) at (8.4+\disttopgraphx,2.3) {};
    
\end{tikzpicture}
	\caption{ED in operation on a perturbed social network among voters. Black solid edges are the true links and dashed red edges represent adversarially perturbed links.}
	\label{fig:sampled}
\end{figure}

In the previous section, the nodes were involved in $I$ different relations, with each slab of our tensor graph $\underline{\mathbf{A}}$ representing one of those relations. In this section, the TGCN architecture is leveraged to robustify popular \emph{single-graph} GCNs. Consider that the nodes are involved in a \emph{single} relation represented by the graph $\bar{\mathcal{G}}$ that does not necessarily represent the true graph, but an approximate (nominal) version of it. This shows up for example, in applications involving random graph models \cite{newman2018networks,bollobas2001random,harris2013introduction}, where $\bar{\mathcal{G}}$ is 
one of {the} multiple possible realizations. Similarly, this model also fits adversarial settings, where the links of the nominal graph $\bar{\mathcal{G}}$ are corrupted by a foe (see Fig. \ref{fig:sampled} for a more detailed illustration of this setup). 

Our approach here is to use $\bar{\mathcal{G}}=(\mathcal{V},\mathbf{A})$ to generate a set of $I$ candidate graphs $\{\mathcal{G}_{i}=(\mathcal{V},\mathbf{A}_{i})\}_{i=1}^I$, whose adjacency matrices form the tensor $\underline{\mathbf{A}}$. Clearly, this approach can also be used for multi-relational graphs, generating multiple candidate graphs per relation. The ensuing subsections elaborate on three scenarios of interest.

\subsection{Robustness to the graph topology identification method}
In applications dealing with communications, power, and transportation systems, the network connectivity may be explicitly known. In several other {settings, however, the graph is implicit and} must be learned from observed data. Several methods to infer the topology exist, each relying on a different model that relates the graph with data interdependencies~\cite{giannakis2017tutor}. Since in most applications a ground-truth graph is not available, one faces the challenge of selecting the appropriate graph-aware learning approach. Especially for the setup considered here, the approach  selected (and hence the resultant graph) will have an impact on GCN performance.  

Consider first the $\kappa$-nearest neighbors ($\kappa$-NN) method, the `workhorse' SSL approach that is employed to construct graphs in data mining and machine learning tasks, including regression, classification, collaborative filtering, and similarity search, to list just a few~\cite{dong2011efficient}. Whether nodes ${n}$ and ${n'}$ are linked in $\kappa$-NN depends on a distance metric between their nodal features. For the Euclidean distance, we simply have $d(n,n')=\|\mathbf{x}_{n}-\mathbf{x}_{n'}\|_2^2$. Then, for each node $n$ the distances with respect to all other nodes $n'\neq n$ are ranked and $n$ is connected with the $\kappa$ nodes with the smallest distances $\{d(n,n')\}$. However, selecting the appropriate $\kappa$ and distance metric $d(\cdot,\cdot)$ is often arbitrary, and may not generalize well to unseen data, especially if the learning system operates in an online fashion. Hence, our approach to robustify SSL in that scenario is to consider a tensor graph where each slab corresponds to a graph constructed using a different value of $\kappa$ and (or) distance.

A similar challenge arises in the so-called correlation network methods \cite{giannakis2017tutor}. Here the topology is identified using pairs of feature vectors at nodes $n$ and $n'$, by comparing the sample correlation coefficient $\rho_{nn'}$ to a threshold $\eta$, and asserting the nonzero edge weight as $\rho_{nn'}$ if $|\rho_{nn'}|>\eta$. Selecting $\eta$ depends on the prescribed false-alarm rate, and can compromise the GCN's learning performance. If cause-effect links are of interest, partial correlation coefficients that can be related to the inverse covariance matrix of the nodal features, have well-documented merits, especially when regularized with edge sparsity as in the graphical Lasso method; see e.g., \cite{giannakis2017tutor} and references therein. 

In such cases, our fresh idea is to collect the multiple learned graphs, originating from possibly different methods, as slabs of $\underline{\mathbf{A}}$, and then train our TGCN architecture. Depending on the application at hand, it may be prudent to include in the training a block-sparsity penalty on the coefficients $\underline{\mathbf{R}}$, so that we exploit possibly available prior information on the most appropriate graphs. 

\subsection{Robustness to edge attacks via edge dithering}

The ever-expanding interconnection of social, email, and media service platforms presents an opportunity for adversaries manipulating networked data to launch malicious attacks~\cite{zugner18adv,goodfellow2014explaining,aggarwal2015outlier}. Perturbed  edges modify the graph neighborhoods, which can markedly degrade the learning performance of GCNs. With reference to Fig.~\ref{fig:multilayer}, several edges of the voting network can be adversarially manipulated so that the voters are steered toward a specific direction. This section explains how TGCN can deal with learning applications, where graph edges have been adversarially perturbed. 

To this end, we introduce our so-termed edge dithering (ED) module, which for the given nominal graph, creates a new graph by randomly adding/removing links with the aim to restore a node's initial graph neighborhood. Dithering in visual and audio applications, refers to the intentional injection of noise so that the quantization error is converted to random noise, which is visually more desirable~\cite{ulichney1988dithering}. We envision using our TGCN with each slab of the tensor  $\underline{\mathbf{A}}$ corresponding to a graph that has been obtained after dithering some of the edges in the nominal (and potentially compromised) graph $\bar{\mathcal{G}}$. 	

Mathematically, given the (perturbed) graph $\bar{\mathcal{G}}=(\mathcal{V},\bar{\mathbf{A}})$, we generate $I$ ED graphs $\{\mathcal{G}_{i}\}_{i=1}^{I}$, with $\mathcal{G}_{i}=(\mathcal{V},\mathbf{A}_{i})$, and with the edges of the auxiliary graph $\mathbf{A}_{i}$ selected randomly as 
\begin{align}
	\label{eq:samplegraph}
	A_{n,n',i}=\left\{
	\begin{array}{ll}
	1& \text{wp.}~~~q_1^{ \delta(\bar{A}_{n,n'}=1)}{(1-q_2)}^{  \delta(\bar{A}_{n,n'}=0)}\\
	0&\text{wp.}~~~q_2^{ \delta(\bar{A}_{n,n'}=0)}{(1-q_1)}^{  \delta(\bar{A}_{n,n'}=1)}
	\end{array} 
	\right. 
	\end{align}
where $\delta(\cdot)$ is the indicator function; while the dithering probabilities are set to $q_1={\rm Pr}(A_{n,n',i}=1|\bar{A}_{n,n'}=1)$, and 
$q_2={\rm Pr}(A_{n,n',i}=0|\bar{A}_{n,n'}=0)$. If $n$ and $n'$ are connected in $\bar{\mathcal{G}}$, the edge connecting $n$ with $n'$ is deleted with probability $1-q_1$. Otherwise, if $n$ and $n'$ are not connected in $\bar{\mathcal{G}}$ i.e. $(\bar{A}_{n,n'}=0)$, an edge between $n$ and $n'$ is inserted with probability $1-q_2$. 

The ED graphs give rise to different neighborhoods $\mathcal{N}_n^{(i)}$, and the role of the ED module is to ensure that the unperturbed neighborhood of each node will be present with high probability in at least one of the $I$ graphs. For clarity, we formalize this intuition in the ensuing remark.  
\begin{myremark}
With high probability, there exists $\mathcal{G}_i$ such that a perturbed edge will be restored to its initial value. This means that there exists an ED graph $i$ such that $A_{n,n',i}=A_{n,n'}$. {Since each} $\mathcal{G}_i$ is independently drawn, it holds that 
		\begin{align}
		{\rm Pr}\big(\Pi_{i=1}^I\delta(A_{n,n',i}=1)\big|\bar{A}_{n,n'}=1,A_{n,n'}=0\big)=q_1^I\nonumber\\
		\nonumber{\rm Pr}\big(\Pi_{i=1}^I\delta(A_{n,n',i}=0)\big|\bar{A}_{n,n'}=0,A_{n,n'}=1\big)=q_2^I.
		\end{align}
\end{myremark}
That is, as $I$ increases, the probability that the true edge weight appears in none of the perturbed graphs, decreases exponentially. By following a similar argument, one can argue that, as $I$ increases, the probability that none of the graphs recovers the original neighborhood structure decreases, so that there exists an ED graph $i$ such that $\mathcal{N}_n^{(i)}=\mathcal{N}_n$. At least as important, since TGCN linearly combines (outputs of) different graphs, it will effectively span the range of graphs that we are able to represent, rendering the overall processing scheme less sensitive to adversarial edge perturbations. Indeed, numerical experiments with adversarial attacks will demonstrate that, even with a small $I$, the use of ED significantly boosts classification performance. The operation of the ED module is illustrated in Fig.~\ref{fig:sampled}. 
	
\subsection{Learning over random graphs} 

Uncertainty is ubiquitous in nature and graphs are no exception. Hence, an important research directions is to develop meaningful and tractable models for random graphs. Such models have originated from the graph-theory community (from the early Erd\H{o}s-R\'enyi models to more-recent low-rank graphon generalizations \cite{bollobas2001random}), but also from the network-science community (e.g., preferential attachment models \cite[Ch. 12-16]{newman2018networks}), and the statistics community (e.g., exponential random graph models \cite{harris2013introduction}). These random graph models provide valuable tools for studying structural features of networks, such as giant and small components, degree distributions, path  lengths, and so forth. They further provide parsimonious parametric models that can be leveraged to solve challenging inference and inverse problems. This is the context of having access to limited graph-related observations such as the induced graph at a subset of nodes, or the mean and variance of certain graph motifs; see e.g., \cite{alon2007network}. In those cases, inferring the full graph can be infeasible, but one can postulate a particular random graph model, and utilize the available observations to infer the parameters that best fit the data. 

It is then natural to ponder whether such random graph models can be employed to learn from an (incomplete) set of graph signals using GCNs. Several alternatives are available, including, for example, implementing a multi-layer GCN with a different realization of the graph per layer~\cite{isufi2017filtering}. Differently, we advocate here to leverage once again our TGCN architecture. Our idea is to draw $I$ realizations of the random graph model, form the $N \times N \times I$ tensor $\underline{\mathbf{A}}$, and train a TGCN. This way, each layer considers not only one, but multiple realizations of the graph. Clearly, if we consider an online setup where GCN layers are associated with time, the proposed model can be related to importance sampling and particle filtering approaches, with each slab of the tensor $\underline{\mathbf{A}}$ representing a different particle of the graph probability space \cite{candy2016bayesian}. This hints at the possibility of developing TGCN schemes for the purpose of nonlinear Bayesian estimation over graphs. While certainly of interest, this will be part of our future research agenda.

\section{Numerical tests}\label{sec:ntest}
This section tests the performance of TGCN in learning from multiple potentially perturbed graphs, and provides tangible answers to the following questions.
\begin{itemize}
		\item[\textbf{Q1}.] How does TGCN compare to state-of-the-art methods for SSL over multi-relational graphs?
		\item[\textbf{Q2}.] How can TGCN leverage topologies learned from multiple graph-learning methods?
		\item[\textbf{Q3}.] How robust is TGCN compared to GCN in the presence of noisy features, noisy edge weights, and random as well as adversarial edge perturbations?
		\item[\textbf{Q4}.] How sensitive is TGCN to the ED parameters, namely $q_1$, $q_2$, and $I$?
\end{itemize}

Unless stated otherwise, we test the proposed TGCN with $R=2$,  $L=3$, $P^{(1)}=64$,
$P^{(2)}=8$, and  $P^{(3)}=K$. The regularization parameters  $\{\mu_1,\mu_2,\lambda\}$ are chosen based on the performance of the TGCN in the validation set of each experiment. For training, an ADAM optimizer with learning rate 0.005 was employed \cite{kingma2015adam}, for 300 epochs\footnote{An epoch is a cycle through all the training examples} with early stopping at 60 epochs\footnote{Training stops if the validation loss does not decrease for 60 epochs}. The 
simulations were run using TensorFlow~\cite{abadi2016tensorflow}, and the code is available 
online\footnote{https://sites.google.com/site/vasioannidispw/github}. 
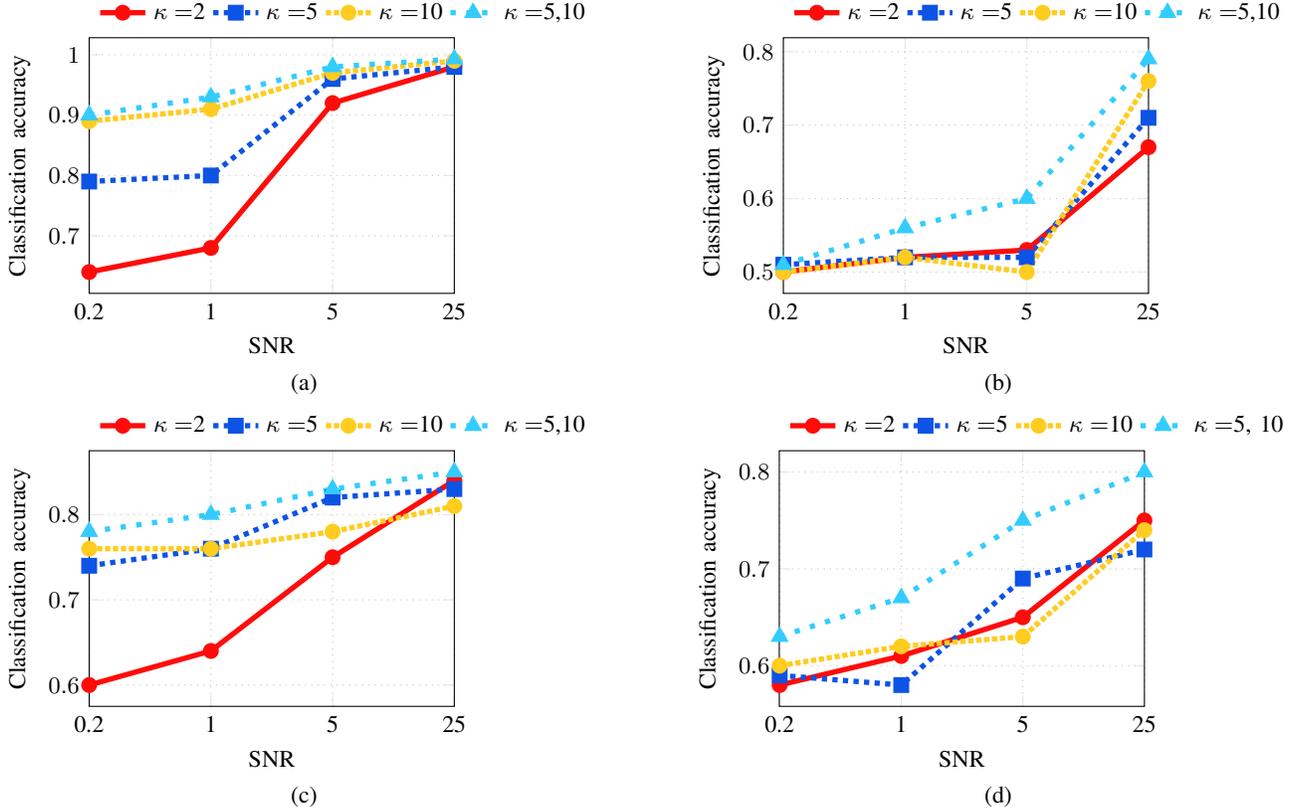
\begin{figure*}[t]
    \begin{subfigure}[b]{0.5\columnwidth}
    \centering{% This file was created by matlab2tikz.
%
%The latest updates can be retrieved from
%  http://www.mathworks.com/matlabcentral/fileexchange/22022-matlab2tikz-matlab2tikz
%where you can also make suggestions and rate matlab2tikz.
%
\begin{tikzpicture}

\begin{axis}[%
width=0.956\mywidth,
height=0.987\myheight,
at={(0\mywidth,0\myheight)},
scale only axis,
xlabel style={font=\color{white!15!black}},
xlabel={SNR},
xmin=0.2,
xmax=25,
xmode=log,xtick={0.2,1,5,25,125},xticklabels={0.2,1,5,25,125},
ylabel style={font=\color{white!15!black}},
ylabel={Classification accuracy},
legend columns=4,
xmajorgrids,
ymajorgrids,
  cycle list name=colorlist,
grid style={dotted},ticklabel style={font=\ticklabelfontsize},
legend style={
	at={(0,1.015)}, 
	anchor=south west, legend cell align=left, align=left,
	draw=none
	% white!15!black
	, font=\legendfontsize},ticklabel style={font=\ticklabelfontsize},label style={font=\mlabelfontsize}
]
\addplot%[color=mycolor1, line width=2.0pt, mark=o, mark options={solid, mycolor1}]
  table[row sep=crcr]{%
0.2	0.64\\
1	0.68\\
5 0.92\\
25	0.98\\
%125	0.99\\
};
\addlegendentry{$\kappa=$2}
\addplot
  table[row sep=crcr]{%
0.2	0.79\\
1	0.80\\
5 0.96\\
25	0.98\\
%125	0.99\\
};
\addlegendentry{$\kappa=$5}
\addplot%[color=mycolor2, line width=2.0pt, dashed, mark=square, mark options={solid, mycolor2}]
  table[row sep=crcr]{%
0.2	0.89\\
1	0.91\\
5 0.97\\
25	0.99\\
%125	0.998\\
};
\addlegendentry{$\kappa=$10}
\addplot
%[color=mycolor2, line width=2.0pt, dashed, mark=square, mark options={solid, mycolor2}]
  table[row sep=crcr]{%
0.2	0.90\\
1	0.93\\
5 0.98\\
25	0.993\\
%125	0.999\\
};
\addlegendentry{$\kappa=$5,10}
\end{axis}
\end{tikzpicture}%
}
    \vspace{-.1cm}
	\caption{ }
    \end{subfigure}%
    ~ 
    \begin{subfigure}[b]{0.5\columnwidth}
    \centering{% This file was created by matlab2tikz.
%
%The latest updates can be retrieved from
%  http://www.mathworks.com/matlabcentral/fileexchange/22022-matlab2tikz-matlab2tikz
%where you can also make suggestions and rate matlab2tikz.
%
\begin{tikzpicture}

\begin{axis}[%
width=0.956\mywidth,
height=0.987\myheight,
at={(0\mywidth,0\myheight)},
scale only axis,
xlabel style={font=\color{white!15!black}},
xlabel={SNR},
xmin=0.2,
xmax=25,
xmode=log,xtick={0.2,1,5,25,125},xticklabels={0.2,1,5,25,125},
ylabel style={font=\color{white!15!black}},
ylabel={Classification accuracy},
legend columns=4,
xmajorgrids,
ymajorgrids,
  cycle list name=colorlist,
grid style={dotted},ticklabel style={font=\ticklabelfontsize},
legend style={
	at={(0,1.015)}, 
	anchor=south west, legend cell align=left, align=left,
	draw=none
	% white!15!black
	, font=\legendfontsize},ticklabel style={font=\ticklabelfontsize},label style={font=\mlabelfontsize}
]
\addplot%[color=mycolor1, line width=2.0pt, mark=o, mark options={solid, mycolor1}]
  table[row sep=crcr]{%
0.2	0.50\\
1	0.52\\
5 0.53\\
25	0.67\\
125	0.98\\
};
\addlegendentry{$\kappa=$2}
\addplot
  table[row sep=crcr]{%
0.2	0.51\\
1	0.52\\
5 0.52\\
25	0.71\\
125	0.99\\
};
\addlegendentry{$\kappa=$5}
\addplot%[color=mycolor2, line width=2.0pt, dashed, mark=square, mark options={solid, mycolor2}]
  table[row sep=crcr]{%
0.2	0.50\\
1	0.52\\
5 0.50\\
25	0.76\\
125	0.99\\
};
\addlegendentry{$\kappa=$10}
\addplot
%[color=mycolor2, line width=2.0pt, dashed, mark=square, mark options={solid, mycolor2}]
  table[row sep=crcr]{%
0.2	0.51\\
1	0.56\\
5 0.60\\
25	0.79\\
125	0.999\\
};
\addlegendentry{$\kappa=$5,10}
\end{axis}
\end{tikzpicture}%
}
	\vspace{-.1cm}
	\caption{ }
    \end{subfigure}\\%
%	\caption{	}
%\label{fig:synthetic}
%\end{figure*}
%\begin{figure*}[!htb]
\vspace{.1cm}
    \begin{subfigure}[b]{0.5\columnwidth}
    \centering{% This file was created by matlab2tikz.
%
%The latest updates can be retrieved from
%  http://www.mathworks.com/matlabcentral/fileexchange/22022-matlab2tikz-matlab2tikz
%where you can also make suggestions and rate matlab2tikz.
%
\begin{tikzpicture}

\begin{axis}[%
width=0.956\mywidth,
height=0.987\myheight,
at={(0\mywidth,0\myheight)},
scale only axis,
xlabel style={font=\color{white!15!black}},
xlabel={SNR},
xmin=0.2,
xmax=25,
xmode=log,xtick={0.2,1,5,25,125},xticklabels={0.2,1,5,25,125},
ylabel style={font=\color{white!15!black}},
ylabel={Classification accuracy},
legend columns=4,
xmajorgrids,
ymajorgrids,
  cycle list name=colorlist,
grid style={dotted},
legend style={
	at={(0,1.015)}, 
	anchor=south west, legend cell align=left, align=left,
	draw=none
	% white!15!black
	, font=\legendfontsize},ticklabel style={font=\ticklabelfontsize},label style={font=\mlabelfontsize}
]
\addplot%[color=mycolor1, line width=2.0pt, mark=o, mark options={solid, mycolor1}]
  table[row sep=crcr]{%
0.2	0.6\\
1	0.64\\
5 0.75\\
25	0.84\\
};
\addlegendentry{$\kappa=$2}
\addplot
  table[row sep=crcr]{%
0.2	0.74\\
1	0.76\\
5 0.82\\
25	0.83\\
};
\addlegendentry{$\kappa=$5}

\addplot%[color=mycolor2, line width=2.0pt, dashed, mark=square, mark options={solid, mycolor2}]
  table[row sep=crcr]{%
0.2	0.76\\
1	0.76\\
5 0.78\\
25	0.81\\
};
\addlegendentry{$\kappa=$10}
\addplot
%[color=mycolor2, line width=2.0pt, dashed, mark=square, mark options={solid, mycolor2}]
  table[row sep=crcr]{%
0.2	0.78\\
1	0.80\\
5 0.83\\
25	0.85\\
};
\addlegendentry{$\kappa=$5,10}
\end{axis}
\end{tikzpicture}%
}
\vspace{-.1cm}
\caption{ }
    \end{subfigure}%
    ~ 
    \begin{subfigure}[b]{0.5\columnwidth}
    \centering{% This file was created by matlab2tikz.
%
%The latest updates can be retrieved from
%  http://www.mathworks.com/matlabcentral/fileexchange/22022-matlab2tikz-matlab2tikz
%where you can also make suggestions and rate matlab2tikz.
%
\begin{tikzpicture}

\begin{axis}[%
width=0.956\mywidth,
height=0.987\myheight,
at={(0\mywidth,0\myheight)},
scale only axis,
xlabel style={font=\xlabelfontsize},
xlabel={SNR},
xmin=0.199,
xmax=25.001,
xmode=log,
xtick={0.2,1,5,25,125},
xticklabels={0.2,1,5,25,125},label style={font=\tiny},
ylabel style={font=\xlabelfontsize},
ylabel={Classification accuracy},
legend columns=4,
xmajorgrids,
ymajorgrids,
  cycle list name=colorlist,
grid style={dotted},ticklabel style={font=\tiny},
legend style={
		at={(0,1.015)}, 
	anchor=south west, legend cell align=left, align=left,
	draw=none
	% white!15!black
	, font=\legendfontsize},ticklabel style={font=\ticklabelfontsize},label style={font=\mlabelfontsize}
]
\addplot%[color=mycolor1, line width=2.0pt, mark=o, mark options={solid, mycolor1}]
  table[row sep=crcr]{%
0.2	0.58\\
1	0.61\\
5 0.65\\
25	0.75\\
125	0.84\\
};
\addlegendentry{$\kappa=$2}
\addplot
  table[row sep=crcr]{%
0.2	0.59\\
1	0.58\\
5 0.69\\
25	0.72\\
125	0.84\\
};
\addlegendentry{$\kappa=$5}
\addplot%[color=mycolor2, line width=2.0pt, dashed, mark=square, mark options={solid, mycolor2}]
  table[row sep=crcr]{%
0.2	0.60\\
1	0.62\\
5 0.63\\
25	0.74\\
125	0.85\\
};
\addlegendentry{$\kappa=$10}
\addplot
%[color=mycolor2, line width=2.0pt, dashed, mark=square, mark options={solid, mycolor2}]
  table[row sep=crcr]{%
0.2	0.63\\
1	0.67\\
5 0.75\\
25	0.80\\
125	0.88\\
};
\addlegendentry{$\kappa=$5, 10}
\end{axis}

\end{tikzpicture}%
}
	\vspace{-.1cm}
	\caption{ }
    \end{subfigure}%
	\caption{Classification accuracy on the synthetic (a)-(b) and ionosphere (c)-(d) graphs described in Sec. \ref{Sec:NumericalSNR} as the noise level in the features [cf. \eqref{eq:featpertub}] or in the links [\eqref{eq:toppertub}] varies. Panels (a) and (c) show the classification accuracy for noisy features while panels (b) and (d) show the same metric as the power of the noise added to the graph links  varies. 
	}
\label{fig:robust}
\end{figure*}
\subsection{SSL using multiple learned graphs}\label{Sec:NumericalSNR}
This section reports the performance of the proposed architecture when multiple learned graphs are employed and data are corrupted by noise. When topologies and feature vectors are noisy, the observed $\underline{\mathbf{A}}$ and $\mathbf{X}$ is modeled as 
\begin{align}
    \label{eq:toppertub}
    \underline{\mathbf{A}}=&\underline{\mathbf{A}}_{tr}+\underline{\mathbf{O}}_\mathbf{A}\\
     \mathbf{X}=&\mathbf{X}_{tr}+\underline{\mathbf{O}}_\mathbf{X}\label{eq:featpertub}.
\end{align}
where $\underline{\mathbf{A}}_{tr}$ and $\mathbf{X}_{tr}$ represent the \textit{true} (nominal) topology and features, while $\underline{\mathbf{O}}_\mathbf{A}$ and $\underline{\mathbf{O}}_\mathbf{X}$ denote the corresponding additive perturbations (outliers). We draw  $\underline{\mathbf{O}}_\mathbf{A}$ and $\underline{\mathbf{O}}_\mathbf{X}$  from a zero-mean uncorrelated multivariate Gaussian distribution with specified signal to noise ratio (SNR). The robustness of our method is tested in two datasets: i)  A synthetic dataset of ${N}=1,000$ samples that belong to $K=2$ classes generated as $\mathbf{x}_{{n}}\in\rfield^{ F\times1}\sim\mathcal{N}(\mathbf{m}_x,0.4\mathbf{I})$ for
${n}=1,\ldots,1,000$, with $ F=10$ and the mean vector $\mathbf{m}_x\in \rfield^{ F\times1}$ being all zeros for the first class and all ones for the second class.  ii) The ionosphere dataset, which contains ${N}=351$ data with $F=34$ features that belong to $K=2$ classes \cite{Dua:2017}. We generate $\kappa$-NN graphs by varying $\kappa$, and observe  $|\mathcal{M}|=200$ and $|\mathcal{M}|=50$ nodes uniformly at random.

With this simulation setup, we test the different TGCNs in SSL for increasing SNR values (Figs. \ref{fig:robust}a, \ref{fig:robust}b, \ref{fig:robust}c, \ref{fig:robust}d). We deduce from the classification performance of our method in Fig. \ref{fig:robust} that multiple graphs lead to learning more robust representations of the data, demonstrating the merits of the proposed TGCN architecture. 

\subsection{Robustness of TGCNs to random graph perturbations}\label{Sec:NumericalCitationED}
For this experiment, the novel ED module and TGCN architecture are used to account for perturbations on the graph edges.

In this case, the experiments are run using three citation network datasets from~\cite{sen2008collective}. The adjacency matrix of the citation graph is $\mathbf{A}$, its nodes correspond to different documents from the same scientific category, and $A_{nn'}=1$ implies that paper $n$ cites paper $n'$. Each document ${n}$ is associated with a label $y_{n}$ that indicates the document's subcategory. ``Cora'' contains papers related to machine learning, ``Citeseer'' includes papers related to computer and information science, while ``Pubmed''  contains biomedical-related papers, see also Table \ref{tab:citation}. To facilitate comparison, we reproduce the same experimental setup as in  \cite{kipf2016semi}, i.e., the same split of the data in training, validation, and testing subsets.  For this experiment, the perturbed graph $\bar{\mathbf{A}}$ is generated by inserting new edges in the original graphs between a random pair of nodes $n,n'$ that are not connected in $\mathbf{A}$, meaning $A_{n,n'}=0$. This can represent, for example, documents that should have been cited, but the authors missed. The added edges can be regarded as drawn from Bernoulli's distribution. TGCN utilizes the multiple graphs generated via the ED module with $I=10$ samples, $q_1=0.9$, and $q_2=1$, since no edge is deleted in $\bar{\mathbf{A}}$.

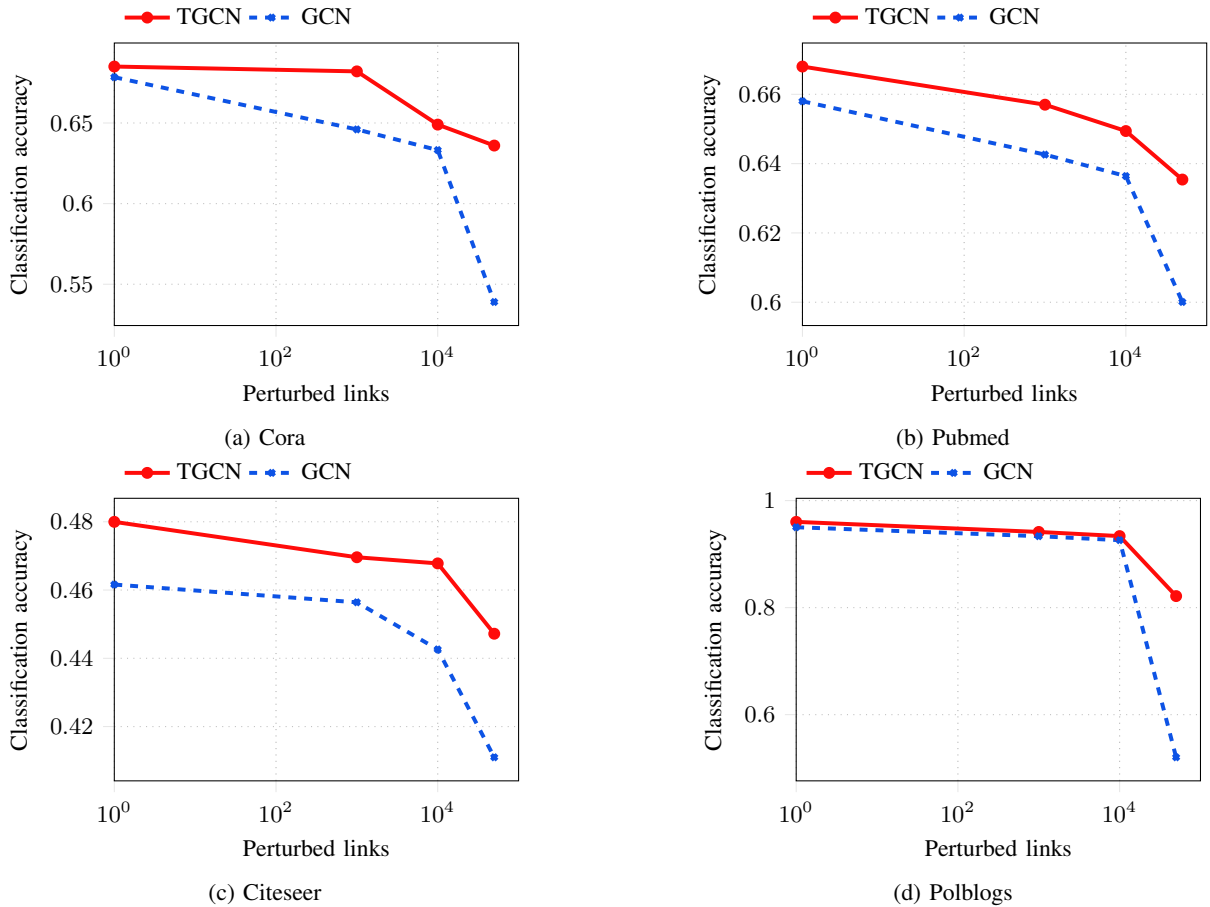
\begin{figure*}
		%\centering
		\begin{subfigure}[b]{0.5\columnwidth}
		\centering% This file was created by matplotlib2tikz v0.6.18.
\begin{tikzpicture}

\definecolor{color0}{rgb}{1,0.647058823529412,0}
\definecolor{color1}{rgb}{1,1,0}
\definecolor{color2}{rgb}{0.501960784313725,0,0.501960784313725}

\begin{axis}[width=0.956\mywidths,
height=0.987\myheights,
at={(0\mywidth,0\myheight)},
%legend entries={{custom-nonlin-invariant},{custom-lin-invariant},{custom-nonlin-raw},{diff-nonlin-invariant},{diff-nonlin-raw},{monicCubic-nonlin-invariant},{monicCubic-nonlin-raw},
%{pointnet}},
legend style={draw=white!80.0!black},
tick align=outside,
tick pos=left,scale only axis,
x grid style={white!69.01960784313725!black},
xlabel={Perturbed links},
xmin=1, xmax=100000, xmode=log,
xtick={1,100,10000},
%xticklabel={\pgfmathparse{\tick*100/10312}\pgfmathprintnumber{\pgfmathresult}\%},
%y grid style={white!69.01960784313725!black},
ylabel={Classification accuracy},
xmajorgrids,
ymajorgrids,ticklabel style={font=\tiny},
grid style={dotted},
legend columns=2,
legend style={
	at={(0,1.015)}, 
	anchor=south west, legend cell align=left, align=left, draw=none
	% white!15!black
	,font=\legendfontsize},ticklabel style={font=\ticklabelfontsize},label style={font=\mlabelfontsize}]

\addplot [line width=\mylinewidth,AGNN, mark=*, mark size=\markwidth, mark options={solid}]
table [row sep=\\]{%
1 0.685\\
1000 0.6820003390312195\\%'max,config:features=0,model=agrcn_edgesamplelearn_rate=0.05,smooth_reg=1e-06,hidden_units1=32hidden_units2=0epochs=250,dropout_rate=0.1,weight_decay=1e-06early_stopping=20,neighbor_list=[],sampl_nbr=2,sample_pct=0.99,max_degree=3,sparse_reg=1e-06': 0.6820003390312195}
10000   0.6490002512931824\\		%config:features=0,model=agrcn_edgesamplelearn_rate=0.05,smooth_reg=1e-06,hidden_units1=32hidden_units2=0epochs=250,dropout_rate=0.1,weight_decay=1e-06early_stopping=20,neighbor_list=[],sampl_nbr=2,sample_pct=0.98,max_degree=3,sparse_reg=1e-06': 0.6490002512931824}
50000 0.636000764369964\\%'max,config:features=0,model=agrcn_edgesamplelearn_rate=0.05,smooth_reg=1e-06,hidden_units1=32hidden_units2=0epochs=250,dropout_rate=0.1,weight_decay=1e-06early_stopping=20,neighbor_list=[],sampl_nbr=2,sample_pct=0.99,max_degree=3,sparse_reg=1e-06': 0.5936000764369964
};\addlegendentry{TGCN}
\addplot [mark size=\markwidth, line width=\mylinewidth, GCN, dashed, mark=x,  mark options={solid}]
table [row sep=\\]{%
1 0.6784003496170044\\%'max,config:features=0,model=gcnlearn_rate=0.05,smooth_reg=1e-06,hidden_units1=64hidden_units2=0epochs=250,dropout_rate=0.1,weight_decay=1e-06early_stopping=100,neighbor_list=[],sampl_nbr=2,sample_pct=0.98,max_degree=3,sparse_reg=1e-06': 0.6784003496170044
1000 0.6460002422332763\\%ax,config:features=0,model=gcnlearn_rate=0.05,smooth_reg=1e-06,hidden_units1=64hidden_units2=8epochs=250,dropout_rate=0.7,weight_decay=1e-06early_stopping=20,neighbor_list=[],sampl_nbr=2,sample_pct=0.98,max_degree=3,sparse_reg=1e-06': 0.6460002422332763}
10000  0.633200216293335			\\%,config:features=0,model=gcnlearn_rate=0.05,smooth_reg=0.01,hidden_units1=32hidden_units2=8epochs=250,dropout_rate=0.1,weight_decay=1e-06early_stopping=20,neighbor_list=[],sampl_nbr=2,sample_pct=0.98,max_degree=3,sparse_reg=1e-06': 0.633200216293335
50000 0.5389999330043793\\%'max,config:features=0,model=gcnlearn_rate=0.05,smooth_reg=1e-06,hidden_units1=32hidden_units2=0epochs=250,dropout_rate=0.7,weight_decay=1e-06early_stopping=20,neighbor_list=[],sampl_nbr=2,sample_pct=0.98,max_degree=3,sparse_reg=1e-06': 0.5389999330043793}
};
\addlegendentry{GCN}
\end{axis}
\end{tikzpicture}
		\caption{Cora}
		\end{subfigure}~\begin{subfigure}[b]{0.5\columnwidth}
		\centering% This file was created by matplotlib2tikz v0.6.18.
\begin{tikzpicture}

\begin{axis}[width=0.956\mywidths,
height=0.987\myheights,
at={(0\mywidth,0\myheight)},
legend style={draw=white!80.0!black},
tick align=outside,scale only axis,
tick pos=left,
x grid style={white!69.01960784313725!black},
xlabel={Perturbed links},
xmin=1, xmax=100000, xmode=log,
xtick={1,100,10000},%xticklabel={\pgfmathparse{\tick*100/10312}\pgfmathprintnumber{\pgfmathresult}\%},
%y grid style={white!69.01960784313725!black},
ylabel={Classification accuracy},
xmajorgrids,
ymajorgrids,
grid style={dotted},
legend columns=2,ticklabel style={font=\tiny},
legend style={
	at={(0.0,1.015)}, 
	anchor=south west, legend cell align=left, align=left, draw=none
	% white!15!black
	,font=\legendfontsize},ticklabel style={font=\ticklabelfontsize},label style={font=\mlabelfontsize}]

\addplot [line width=\mylinewidth,AGNN, mark=*, mark size=\markwidth, mark options={solid}]
table [row sep=\\]{%
1 0.668008856773377\\
1000   0.6570007920265197\\		%'max,config:features=0,model=gcnlearn_rate=0.05,smooth_reg=1e-06,hidden_units1=32hidden_units2=8epochs=250,dropout_rate=0.1,weight_decay=1e-06early_stopping=20,neighbor_list=[],sampl_nbr=2,sample_pct=0.98,max_degree=3,sparse_reg=1e-06': 0.6570007920265197}
10000 0.6494006896018982\\%'max,config:features=0,model=gcnlearn_rate=0.05,smooth_reg=1e-06,hidden_units1=64hidden_units2=0epochs=250,dropout_rate=0.1,weight_decay=1e-06early_stopping=20,neighbor_list=[],sampl_nbr=2,sample_pct=0.98,max_degree=3,sparse_reg=1e-06': 0.6494006896018982}
50000 0.6354006671905517\\%'max,config:features=0,model=gcnlearn_rate=0.05,smooth_reg=1e-06,hidden_units1=64hidden_units2=0epochs=250,dropout_rate=0.1,weight_decay=1e-06early_stopping=20,neighbor_list=[],sampl_nbr=2,sample_pct=0.98,max_degree=3,sparse_reg=1e-06': 0.6154006671905517}
};\addlegendentry{TGCN}

\addplot [mark size=\markwidth, line width=\mylinewidth, GCN, dashed, mark=x,  mark options={solid}]
table [row sep=\\]{%\
1 0.6580008029937744\\
1000 0.6426008152961731\\%'max,config:features=0,model=gcnlearn_rate=0.05,smooth_reg=0.0001,hidden_units1=32hidden_units2=0epochs=250,dropout_rate=0.1,weight_decay=1e-06early_stopping=20,neighbor_list=[],sampl_nbr=10,sample_pct=0.99,max_degree=3,sparse_reg=0.0001': 0.6426008152961731}
10000  0.6364007735252381			\\%'max,config:features=0,model=gcnlearn_rate=0.05,smooth_reg=1e-06,hidden_units1=32hidden_units2=0epochs=250,dropout_rate=0.1,weight_decay=1e-06early_stopping=20,neighbor_list=[],sampl_nbr=2,sample_pct=0.98,max_degree=3,sparse_reg=0.0001': 0.6364007735252381}
50000 0.6000762462616\\%'max,config:features=0,model=gcnlearn_rate=0.05,smooth_reg=0.0001,hidden_units1=64hidden_units2=0epochs=250,dropout_rate=0.1,weight_decay=1e-06early_stopping=20,neighbor_list=[],sampl_nbr=5,sample_pct=0.98,max_degree=3,sparse_reg=0.01': 0.618000762462616}
};\addlegendentry{GCN}

\end{axis}
\end{tikzpicture}
		\caption{Pubmed}
		\end{subfigure}
		
		\begin{subfigure}[b]{0.5\columnwidth}
		\centering% This file was created by matplotlib2tikz v0.6.18.
\begin{tikzpicture}

\definecolor{color0}{rgb}{1,0.647058823529412,0}
\definecolor{color1}{rgb}{1,1,0}
\definecolor{color2}{rgb}{0.501960784313725,0,0.501960784313725}

\begin{axis}[width=0.956\mywidths,
height=0.987\myheights,
at={(0\mywidth,0\myheight)},
legend style={draw=white!80.0!black},
tick align=outside,
tick pos=left,scale only axis,
x grid style={white!69.01960784313725!black},
xlabel={Perturbed links},
xmin=1, xmax=100000, xmode=log,ticklabel style={font=\tiny},
xtick={1,100,10000},%xticklabel={\pgfmathparse{\tick*100/10312}\pgfmathprintnumber{\pgfmathresult}\%},
%y grid style={white!69.01960784313725!black},
ylabel={Classification accuracy},
xmajorgrids,
ymajorgrids,
grid style={dotted},
legend columns=2,
legend style={
	at={(0,1.015)}, 
	anchor=south west, legend cell align=left, align=left, draw=none
	% white!15!black
,font=\legendfontsize},ticklabel style={font=\ticklabelfontsize},label style={font=\mlabelfontsize}]

\addplot [line width=\mylinewidth,AGNN, mark=*, mark size=\markwidth, mark options={solid}]
table [row sep=\\]{%
1 0.479973134994507\\
1000   0.4695997323989868\\		%'max,config:features=0,model=gcnlearn_rate=0.05,smooth_reg=0.0001,hidden_units1=32hidden_units2=0epochs=250,dropout_rate=0.1,weight_decay=1e-06early_stopping=20,neighbor_list=[],sampl_nbr=5,sample_pct=0.98,max_degree=3,sparse_reg=1e-06': 0.4695997323989868}
10000 0.467799711227417\\%'max,config:features=0,model=gcnlearn_rate=0.05,smooth_reg=0.01,hidden_units1=64hidden_units2=0epochs=250,dropout_rate=0.7,weight_decay=1e-06early_stopping=20,neighbor_list=[],sampl_nbr=2,sample_pct=0.98,max_degree=3,sparse_reg=0.0001': 0.467799711227417}
50000 0.44719977498054507\\%'max,config:features=0,model=gcnlearn_rate=0.05,smooth_reg=0.0001,hidden_units1=32hidden_units2=8epochs=250,dropout_rate=0.1,weight_decay=1e-06early_stopping=20,neighbor_list=[],sampl_nbr=10,sample_pct=0.99,max_degree=3,sparse_reg=1e-06': 0.42719977498054507}
};\addlegendentry{TGCN}

\addplot [mark size=\markwidth, line width=\mylinewidth, GCN, dashed, mark=x,  mark options={solid}]
table [row sep=\\]{%
1 0.461599725484848\\
1000 0.4563997304439545\\%'max,config:features=0,model=gcnlearn_rate=0.05,smooth_reg=1e-06,hidden_units1=32hidden_units2=8epochs=250,dropout_rate=0.1,weight_decay=1e-06early_stopping=20,neighbor_list=[],sampl_nbr=2,sample_pct=0.98,max_degree=3,sparse_reg=1e-06': 0.4563997304439545}
10000  0.4425997519493103			\\%'max,config:features=0,model=gcnlearn_rate=0.05,smooth_reg=1e-06,hidden_units1=32hidden_units2=0epochs=250,dropout_rate=0.7,weight_decay=1e-06early_stopping=20,neighbor_list=[],sampl_nbr=2,sample_pct=0.98,max_degree=3,sparse_reg=1e-06': 0.4425997519493103}
50000 0.4109997868537903\\%'max,config:features=0,model=gcnlearn_rate=0.05,smooth_reg=1e-06,hidden_units1=32hidden_units2=0epochs=250,dropout_rate=0.7,weight_decay=1e-06early_stopping=20,neighbor_list=[],sampl_nbr=2,sample_pct=0.98,max_degree=3,sparse_reg=1e-06': 0.4109997868537903}
};\addlegendentry{GCN}

\end{axis}
\end{tikzpicture}
		\caption{Citeseer}
	\end{subfigure}~\begin{subfigure}[b]{0.5\columnwidth}
		\centering% This file was created by matplotlib2tikz v0.6.18.
\begin{tikzpicture}

\definecolor{color0}{rgb}{1,0.647058823529412,0}
\definecolor{color1}{rgb}{1,1,0}
\definecolor{color2}{rgb}{0.501960784313725,0,0.501960784313725}

\begin{axis}[width=0.956\mywidths,
height=0.987\myheights,
at={(0\mywidth,0\myheight)},
%legend entries={{custom-nonlin-invariant},{custom-lin-invariant},{custom-nonlin-raw},{diff-nonlin-invariant},{diff-nonlin-raw},{monicCubic-nonlin-invariant},{monicCubic-nonlin-raw},
%{pointnet}},
legend style={draw=white!80.0!black},
tick align=outside,
tick pos=left,scale only axis,
x grid style={white!69.01960784313725!black},
xlabel={Perturbed links},
xmin=1, xmax=100000, xmode=log,
xtick={1,100,10000},
%xticklabel={\pgfmathparse{\tick*100/10312}\pgfmathprintnumber{\pgfmathresult}\%},
%y grid style={white!69.01960784313725!black},
ylabel={Classification accuracy},
xmajorgrids,
ymajorgrids,
grid style={dotted},
legend columns=2,ticklabel style={font=\tiny},
legend style={
	at={(0,1.015)}, 
	anchor=south west, legend cell align=left, align=left, draw=none
	% white!15!black
,font=\legendfontsize},ticklabel style={font=\ticklabelfontsize},label style={font=\mlabelfontsize}]

\addplot [line width=\mylinewidth,AGNN, mark=*, mark size=\markwidth, mark options={solid}]
table [row sep=\\]{%
1 0.9602509498596191\\
1000 0.9414224982261657\\%'max,config:features=0,model=agrcn_edgesamplelearn_rate=0.05,smooth_reg=1e-06,hidden_units1=32hidden_units2=0epochs=250,dropout_rate=0.1,weight_decay=1e-06early_stopping=20,neighbor_list=[],sampl_nbr=2,sample_pct=0.99,max_degree=3,sparse_reg=1e-06': 0.6820003390312195}
10000   0.9336819410324096\\		%config:features=0,model=agrcn_edgesamplepqlearn_rate=0.05,smooth_reg=1e-08,hidden_units1=64hidden_units2=8epochs=250,dropout_rate=0.7,weight_decay=0.1early_stopping=80,neighbor_list=[],sampl_nbr=10,sample_pct=0.0,p=0.99,q=0.0,max_degree=3,sparse_reg=1e-08:=0.9414224982261657
50000 0.8215480446815491\\%'config:features=0,model=agrcn_edgesamplepqlearn_rate=0.05,smooth_reg=1e-08,hidden_units1=64hidden_units2=8epochs=250,dropout_rate=0.7,weight_decay=1e-08early_stopping=80,neighbor_list=[],sampl_nbr=10,sample_pct=0.0,p=0.99,q=0.0,max_degree=3,sparse_reg=1e-08:=0.8215480446815491
};\addlegendentry{TGCN}
\addplot [mark size=\markwidth, line width=\mylinewidth, GCN, dashed, mark=x,  mark options={solid}]
table [row sep=\\]{%
1 0.950209128856659\\%''max,config:features=0,model=gcnlearn_rate=0.05,smooth_reg=1e-06,hidden_units1=64hidden_units2=8epochs=250,dropout_rate=0.1,weight_decay=0.1early_stopping=100,neighbor_list=[],sampl_nbr=2,sample_pct=0.98,max_degree=3,sparse_reg=1e-06': 0.9383663296699524}
1000 0.9334726691246032\\%ax,config:features=0,model=gcnlearn_rate=0.05,smooth_reg=1e-06,hidden_units1=64hidden_units2=8epochs=250,dropout_rate=0.7,weight_decay=1e-06early_stopping=20,neighbor_list=[],sampl_nbr=2,sample_pct=0.98,max_degree=3,sparse_reg=1e-06': 0.6460002422332763}
10000  0.9260250258445739			\\%,config:features=0,model=gcnlearn_rate=0.05,smooth_reg=0.01,hidden_units1=32hidden_units2=8epochs=250,dropout_rate=0.1,weight_decay=1e-06early_stopping=20,neighbor_list=[],sampl_nbr=2,sample_pct=0.98,max_degree=3,sparse_reg=1e-06': 0.633200216293335
50000 0.5202928304672241\\%'max,config:features=0,model=gcnlearn_rate=0.05,smooth_reg=1e-06,hidden_units1=32hidden_units2=0epochs=250,dropout_rate=0.7,weight_decay=1e-06early_stopping=20,neighbor_list=[],sampl_nbr=2,sample_pct=0.98,max_degree=3,sparse_reg=1e-06': 0.5389999330043793}
};
\addlegendentry{GCN}
\end{axis}
\end{tikzpicture}
		\caption{Polblogs}
		\end{subfigure}
		\caption{Classification accuracy for the setup described in Sec. \ref{Sec:NumericalCitationED} as the number of perturbed edges increases.}
		\label{fig:adrandpert}
\end{figure*}

Fig.~\ref{fig:adrandpert} depicts the classification accuracy of the  GCN~\cite{kipf2016semi} compared to that of the proposed TGCN as the number of perturbed edges is increasing. Clearly, our ED-TGCN is more robust than the standard GCN. Moreover, even when no edges are perturbed, the TGCN outperforms the GCN. This observation may be attributed to noisy links in the original graphs, which hinder classification performance. Furthermore, the SSL performance of the GCN significantly degrades as the number of perturbed edges increases, which suggests that GCN is challenged even by ``random attacks.''
	
\begin{table}[]
\hspace{0cm}
    \centering
    \caption{List of citation graph datasets considered in Secs. \ref{Sec:NumericalCitationED} and \ref{Sec:NumericalCitationED} along with most relevant dimensions.}
     \vspace{0.2cm}
    %\rowcolors[]{1}{white}{gray}
    \begin{tabular}{c c  c }
    \hline
    \textbf {Dataset} & \textbf {Nodes} ${N}$ & \textbf {Classes}
    $K$  \\
  \hline
    \hline
    Cora  & 2,708 & 7  \\
       Citeseer & 3,327 & 6    \\
       Pubmed & 19,717 & 3  \\
        Polblogs & 1,224 & 2\\
            \hline
    \end{tabular}
    
    \label{tab:citation}
\end{table}

{\setlength\extrarowheight{2pt}
\begin{table*}
			\caption{Classification accuracy for the setup described in Sec. \ref{Sec:NumericalCitationED2} as the number of attacked nodes $|\mathcal{T}|$ increases. }
			
			\label{tab:results}
			%\resizebox{0.93\textwidth}{!}{ \renewcommand{\arraystretch}{0.6}
			\centering
			\begin{tabular}{@{}p{2cm}p{2cm}cccccc@{}}
			%\cmidrule[\heavyrulewidth]{1-7}
			    \hline
				\multirow{3}{*}{\vspace*{8pt}\textbf{Dataset}}& \multirow{3}{*}{\vspace*{8pt}\textbf{Method}}&\multicolumn{4}{c}{\textbf{Number of attacked nodes} $|\mathcal{T}|$}\\\cmidrule{3-7}
				& & {\textsc{20}} & {\textsc{30}} & {\textsc{40}} & {\textsc{50}} & {\textsc{60}} 
				\\ 
				%\cmidrule{1-7}
				\hline\hline
				\multirow{2}{*}{\rotatebox{0}{\hspace*{-0pt}{Citeseer}} }
				& \textsc{GCN}  & 60.49 & 56.00 &  61.49 & 56.39 & \textbf{58.99}  \\ 
				& \textsc{TGCN} & \textbf{70.99}& \textbf{56.00} & \textbf{61.49} & \textbf{61.20}  & 58.66 \\     
				\cmidrule{1-7}
				\multirow{2}{*}{\rotatebox{0}{\hspace*{-0pt}{Cora}}} 
				& \textsc{GCN}  & 76.00 & 74.66 &  76.00 & 62.39 & 73.66  \\ 
				& \textsc{TGCN} & \textbf{78.00} & \textbf{82.00} & \textbf{84.00} & \textbf{73.59}  & \textbf{74.99}\\     
				\cmidrule{1-7}
				\multirow{2}{*}{\rotatebox{0}{\hspace*{-0pt}{Pubmed}}} 
				& \textsc{GCN}  & \textbf{74.00} & 71.33 &  68.99 & 66.40 & 69.66  \\ 
				& \textsc{TGCN} & 72.00 & \textbf{75.36} & \textbf{71.44} & \textbf{68.50}  & \textbf{74.43}    \\     
				\cmidrule{1-7}
				\multirow{2}{*}{\rotatebox{0}{\hspace*{-0pt}{Polblogs}}} 
				& \textsc{GCN}  & \textbf{85.03} & 86.00 &  84.99 & 78.79 & 86.91  \\ 
				& \textsc{TGCN} & 84.00 & \textbf{88.00} & \textbf{91.99} & \textbf{78.79}  & \textbf{92.00}   \\   
				%\cmidrule[\heavyrulewidth]{1-7}
				 \hline
			\end{tabular}%}
\end{table*}
}\setlength\extrarowheight{0pt}
	
\subsection{Robustness to adversarial attacks on edges}\label{Sec:NumericalCitationED2}
The original graphs corresponding to Cora, Citeseer, Pubmed, and Polblogs were perturbed using the adversarial setup in~\cite{zugner18adv}, where structural attacks are effected on attributed graphs. These attacks perturb connections adjacent to a set  $\mathcal{T}$ of targeted nodes by adding or deleting edges~\cite{zugner18adv}. Our ED module uses $I=10$ sampled graphs with $q_1=0.9$, and $q_2=0.999$.  For this experiment, 30\% of the nodes are used for training, 30\% for validation, and 40\% for testing. The nodes in $\mathcal{T}$ are in the testing set.
	
Table \ref{tab:results} reports the classification accuracy of the GCN and the proposed TGCN for different numbers of attacked nodes $|\mathcal{T}|$. Different from Fig.~\ref{fig:adrandpert}, where the classification accuracy over the test set is reported, Table \ref{tab:results} reports the classification accuracy over the set of attacked nodes $\mathcal{T}$. It is observed that the proposed TGCN is more robust than GCN under adversarial attacks~\cite{zugner18adv}. This finding justifies the use of the novel ED in conjunction with the TGCN that judiciously selects extracted features originating from non-corrupted neighborhoods.

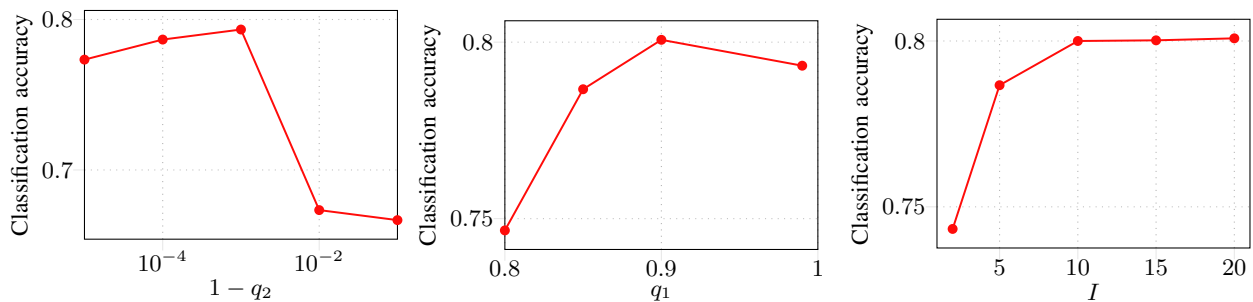
\begin{figure*}[t]
		{% This file was created by matplotlib2tikz v0.6.18.
\begin{tikzpicture}

\begin{axis}[width=0.956\mywidthss,
height=0.987\myheightss,
at={(0\mywidth,0\myheight)},
%legend entries={{custom-nonlin-invariant},{custom-lin-invariant},{custom-nonlin-raw},{diff-nonlin-invariant},{diff-nonlin-raw},{monicCubic-nonlin-invariant},{monicCubic-nonlin-raw},
%{pointnet}},
legend style={draw=white!80.0!black},
tick align=outside,
tick pos=left,scale only axis,
x grid style={white!69.01960784313725!black},
xlabel={$1-q_2$},
xmin=0.00001, xmax=0.1, xmode=log,
xtick={0.0001,0.01},
%xticklabel={\pgfmathparse{\tick*100/10312}\pgfmathprintnumber{\pgfmathresult}\%},
%y grid style={white!69.01960784313725!black},
 ylabel={Classification accuracy},
xmajorgrids,
ymajorgrids,ticklabel style={font=\tiny},
grid style={dotted},label style={font=\tiny},
legend columns=1,
ytick={0.7,0.8},
ticklabel style={font=\tiny,  inner sep=0pt,outer sep=0pt},
legend style={
	at={(0,1.015)}, 
	anchor=south west, legend cell align=left, align=left, draw=none
	% white!15!black
,font=\legendfontsize},ticklabel style={font=\ticklabelfontsize},label style={font=\mlabelfontsize}]

\addplot [line width=0.5*\mylinewidth,AGNN, mark=*, mark size=\markwidth, mark options={solid}]
table [row sep=\\]{%
0.00001 0.7733333826065063\\
0.0001 0.7866667151451111\\%'max,config:features=0,model=agrcn_edgesamplelearn_rate=0.05,smooth_reg=1e-06,hidden_units1=32hidden_units2=0epochs=250,dropout_rate=0.1,weight_decay=1e-06early_stopping=20,neighbor_list=[],sampl_nbr=2,sample_pct=0.99,max_degree=3,sparse_reg=1e-06': 0.6820003390312195}
0.001  0.79333336353302\\		%config:features=0,model=agrcn_edgesamplelearn_rate=0.05,smooth_reg=1e-06,hidden_units1=32hidden_units2=0epochs=250,dropout_rate=0.1,weight_decay=1e-06early_stopping=20,neighbor_list=[],sampl_nbr=2,sample_pct=0.98,max_degree=3,sparse_reg=1e-06': 0.6490002512931824}
0.01 0.6733333706855774\\%'max,config:features=0,model=agrcn_edgesamplelearn_rate=0.05,smooth_reg=1e-06,hidden_units1=32hidden_units2=0epochs=250,dropout_rate=0.1,weight_decay=1e-06early_stopping=20,neighbor_list=[],sampl_nbr=2,sample_pct=0.99,max_degree=3,sparse_reg=1e-06': 0.5936000764369964
0.1 0.6666667103767395\\
};
%\addlegendentry{GCN}
\end{axis}
\end{tikzpicture}}{% This file was created by matplotlib2tikz v0.6.18.
\begin{tikzpicture}

\begin{axis}[width=0.956\mywidthss,
height=0.987\myheightss,
at={(0\mywidth,0\myheight)},
%legend entries={{custom-nonlin-invariant},{custom-lin-invariant},{custom-nonlin-raw},{diff-nonlin-invariant},{diff-nonlin-raw},{monicCubic-nonlin-invariant},{monicCubic-nonlin-raw},
%{pointnet}},
legend style={draw=white!80.0!black},
tick align=outside,
tick pos=left,scale only axis,
x grid style={white!69.01960784313725!black},
xlabel={$q_1$},
xmin=0.8, xmax=1,label style={font=\tiny},
xtick={0.8,0.9,1},
ytick={0.75,0.8},
ticklabel style={font=\tiny, inner sep=0pt,outer sep=0pt},
%xticklabel={\pgfmathparse{\tick*100/10312}\pgfmathprintnumber{\pgfmathresult}\%},
%y grid style={white!69.01960784313725!black},
ylabel={Classification accuracy},
xmajorgrids,
ymajorgrids,ticklabel style={font=\tiny},
grid style={dotted},
legend columns=1,
legend style={
	at={(0,1.015)}, 
	anchor=south west, legend cell align=left, align=left, draw=none
	% white!15!black
,font=\legendfontsize},ticklabel style={font=\ticklabelfontsize},label style={font=\mlabelfontsize}]

\addplot [line width=0.5*\mylinewidth,AGNN, mark=*, mark size=\markwidth, mark options={solid}]
table [row sep=\\]{%
0.8 0.7466667103767395\\
0.85 0.7866667079925537\\
0.9  0.8006667079925537\\	
0.99 0.79333336353302\\
};%\addlegendentry{AGCN}
%\addlegendentry{GCN}
\end{axis}
\end{tikzpicture}} {% This file was created by matplotlib2tikz v0.6.18.
\begin{tikzpicture}

\begin{axis}[width=0.956\mywidthss,
height=0.987\myheightss,
at={(0\mywidth,0\myheight)},
%legend entries={{custom-nonlin-invariant},{custom-lin-invariant},{custom-nonlin-raw},{diff-nonlin-invariant},{diff-nonlin-raw},{monicCubic-nonlin-invariant},{monicCubic-nonlin-raw},
%{pointnet}},
legend style={draw=white!80.0!black},
tick align=outside,
tick pos=left,scale only axis,
x grid style={white!69.01960784313725!black},
xlabel={$I$},
xmin=1, xmax=21,
%xtick={0.0001,0.01},
%xticklabel={\pgfmathparse{\tick*100/10312}\pgfmathprintnumber{\pgfmathresult}\%},
%y grid style={white!69.01960784313725!black},
%ylabel={accuracy},
%ymin=0.356170284628286, ymax=0.858165821954202,
xmajorgrids,
ymajorgrids,ticklabel style={font=\tiny},
grid style={dotted},
ytick={0.75,0.8},
ylabel={Classification accuracy},
ticklabel style={font=\tiny, inner sep=0pt,outer sep=0pt},
legend columns=1,label style={font=\tiny},
legend style={
	at={(0,1.015)}, 
	anchor=south west, legend cell align=left, align=left, draw=none
	% white!15!black
,font=\legendfontsize},ticklabel style={font=\ticklabelfontsize},label style={font=\mlabelfontsize}]

\addplot [line width=0.5*\mylinewidth,AGNN, mark=*, mark size=\markwidth, mark options={solid}]
table [row sep=\\]{%
2 0.7433333945274353\\
5 0.7866666913032532\\
10  0.8000000238418579\\		%config:features=0,model=agrcn_edgesamplelearn_rate=0.05,smooth_reg=1e-06,hidden_units1=32hidden_units2=0epochs=250,dropout_rate=0.1,weight_decay=1e-06early_stopping=20,neighbor_list=[],sampl_nbr=2,sample_pct=0.98,max_degree=3,sparse_reg=1e-06': 0.6490002512931824}
15 0.8002000238418579\\%'max,config:features=0,model=agrcn_edgesamplelearn_rate=0.05,smooth_reg=1e-06,hidden_units1=32hidden_units2=0epochs=250,dropout_rate=0.1,weight_decay=1e-06early_stopping=20,neighbor_list=[],sampl_nbr=2,sample_pct=0.99,max_degree=3,sparse_reg=1e-06': 0.5936000764369964
20 0.8008000238418579\\
};
%\addlegendentry{GCN}
\end{axis}
\end{tikzpicture}}\vspace{-0.0cm}
		\caption{SSL classification accuracy of the TGCN under varying edge creation prob. $q_1$, edge deletion prob. $q_2$, and number of samples $I$.
		}
		\label{fig:robustsens}
\end{figure*}

Fig. \ref{fig:robustsens} showcases the sensitivity of TGCN to varying parameters of the ED module for the experiment in Table \ref{tab:results} with the Cora, and with $|\mathcal{T}|=30$.  Evidently, TGCN's performance is relatively smooth for certain ranges of the parameters. In accordance with Remark 2, notice that even for small $I$ TGCN performance improves considerably. 
	
\subsection{Predicting protein functions}\label{Sec:NumericalProteins}
This section tests the performance of TGCN in \emph{predicting ``protein functions.''}
Protein-to-protein interaction networks relate two proteins via multiple cell-dependent relations that can be modeled using \emph{multi-relational} graphs;  see Fig.\ref{fig:multilayer}.
Protein classification seeks the unknown function of some proteins (nodes) based on the known functionality of a small subset of proteins, and the protein-to-protein networks~\cite{zitnik2017predicting,ioannidis2019camsap}. 

Given a target function $y_n$ that is available on a subset of proteins ${n}\in\mathcal{M}$, known functions on all proteins summarized in $\mathbf{X}$, and
the multi-relational protein networks $\underline{\mathbf{A}}$, the goal is to predict whether the proteins in ${n}\in{\mathcal{V}-\mathcal{M}}$ are associated with the target function 
or not. Hence, the number of target classes is $K=2$. In this setting, $\mathbf{A}_{i}$ represents the protein connectivity in the  ${i}$th cell type, which could be a cerebellum, midbrain, or frontal lobe cell. Table \ref{tab:biodata} summarizes the three datasets used in the following experiments.

\begin{table}[t]
\hspace{0cm}
    \centering
    \caption{List of protein-to-protein interaction datasets considered in Sec. \ref{Sec:NumericalProteins} and their associated dimensions.}
    \vspace{0.2cm}
    %\rowcolors[]{1}{white}{gray}
    \begin{tabular}{c c c c c}
    \hline
    \textbf{Dataset} &  \textbf{Nodes} ${N}$  &  \textbf{Features} $ F$ & \textbf{Relations} ${I}$\\
    \hline\hline
Generic cells  & 4,487 & 502 & 144\\
       Brain cells & 2,702 & 81 & 9  \\
      Circulation cells & 3,385 & 62 & 4\\
 \hline
    \end{tabular}
    \vspace{0.0cm}
    \label{tab:biodata}
\end{table}
We compare TGCN with the GCN in~\cite{kipf2016semi}, which is the single-relational alternative, and Mune~\cite{ye2018robust}, that represents a state-of-the-art diffusion-based
approach for SSL over multi-relational graphs. Since GCN only accounts for a single graph, we select for the GCN the relation $i$ that achieves the best results in the validation set. Furthermore, Mune does not account for feature vectors in the nodes of the graph. For a fair comparison, we employ the TGCN without using the feature vectors, that {is,} $\mathbf{X}=\mathbf{I}_{N}$. Finally, since the classes are heavily unbalanced, we evaluate the performance of the various approaches using the macro F1 score for predicting the protein functions.\footnote{Accurate classifiers achieve macro F1 values close to 1.}
\begin{figure*}
\begin{floatrow}
\ffigbox{%
    \centering
    % This file was created by matlab2tikz.
%
%The latest updates can be retrieved from
%  http://www.mathworks.com/matlabcentral/fileexchange/22022-matlab2tikz-matlab2tikz
%where you can also make suggestions and rate matlab2tikz.
%
\begin{tikzpicture}

\begin{axis}[%
width=0.956\mywidth,
height=0.987\myheight,
at={(0\mywidth,0\myheight)},
scale only axis,
xlabel style={font=\color{white!15!black}},
xlabel={Number of samples},
xmin=50,
xmax=445,
ylabel style={font=\color{white!15!black}},
ylabel={Macro F1 score},
legend columns=2,
xmajorgrids,
ymajorgrids,
grid style={dotted},ticklabel style={font=\ticklabelfontsize},
legend style={
	at={(0,1.015)}, 
	anchor=south west, legend cell align=left, align=left,
	draw=none
	% white!15!black
	, font=\legendfontsize}
]
\addplot [line width=\mylinewidth,AGNN, mark=*, mark size=\markwidth, mark options={solid}]
table [row sep=\\]{%
55 0.69\\
110 0.71\\
220 0.79\\
440  0.86\\
};
\addlegendentry{TGCN}
\addplot [mark size=\markwidth, line width=\mylinewidth, GCN, dashed, mark=x,  mark options={solid}]
table [row sep=\\]{%
55 0.47\\
110 0.48\\
220 0.48\\
440  0.49 \\
};
\addlegendentry{GCN}
\addplot [mark size=\markwidth, line width=\mylinewidth, AGNNnf, dotted, mark=diamond,  mark options={solid}]
table [row sep=\\]{%
55 0.35\\ 
110 0.41\\
220 0.43\\
440  0.41 \\
};
\addlegendentry{TGCN without features}
\addplot [mark size=\markwidth, line width=\mylinewidth, Mune, dashdotted, mark=square,  mark options={solid}]
table [row sep=\\]{%
55  0.14\\
110 0.27\\
220 0.27\\
440  0.32\\
};
\addlegendentry{Mune}
\end{axis}
\end{tikzpicture}%
      \label{fig:bc}
    }{%
  
 \caption{Brain cells}%
}
\ffigbox{%
    \centering
    % This file was created by matlab2tikz.
%
%The latest updates can be retrieved from
%  http://www.mathworks.com/matlabcentral/fileexchange/22022-matlab2tikz-matlab2tikz
%where you can also make suggestions and rate matlab2tikz.
%
\begin{tikzpicture}

\begin{axis}[%
width=0.956\mywidth,
height=0.987\myheight,
at={(0\mywidth,0\myheight)},
scale only axis,
xlabel style={font=\color{white!15!black}},
xlabel={Number of samples},
xmin=50,
xmax=445,
ylabel style={font=\color{white!15!black}},
ylabel={Macro F1 score},
legend columns=2,
xmajorgrids,
ymajorgrids,
grid style={dotted},ticklabel style={font=\ticklabelfontsize},
legend style={
	at={(0,1.015)}, 
	anchor=south west, legend cell align=left, align=left,
	draw=none
	% white!15!black
	, font=\legendfontsize}
]
\addplot [line width=\mylinewidth,AGNN, mark=*, mark size=\markwidth, mark options={solid}]
table [row sep=\\]{%
55 0.69\\
110 0.70\\
220 0.76\\
440  0.77\\
};
\addlegendentry{TGCN}
\addplot [mark size=\markwidth, line width=\mylinewidth, GCN, dashed, mark=x,  mark options={solid}]
table [row sep=\\]{%
55 0.47\\
110 0.48\\
220 0.48\\
440  0.48 \\
};
\addlegendentry{GCN}
\addplot [mark size=\markwidth, line width=\mylinewidth, AGNNnf, dotted, mark=diamond,  mark options={solid}]
table [row sep=\\]{%
55 0.35\\ 
110 0.40\\
220 0.42\\
440  0.41 \\
};
\addlegendentry{TGCN without features}
\addplot [mark size=\markwidth, line width=\mylinewidth, Mune, dashdotted, mark=square,  mark options={solid}]
table [row sep=\\]{%
55  0.13\\
110 0.27\\
220 0.27\\
440  0.28\\
};
\addlegendentry{Mune}
\end{axis}
\end{tikzpicture}%
      \label{fig:cc}
}{%
  
 \caption{Circulation cells}%
}
\end{floatrow}
\end{figure*}

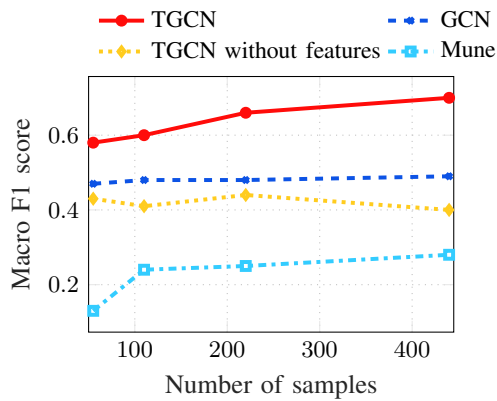
\begin{figure}
    \centering
    % This file was created by matlab2tikz.
%
%The latest updates can be retrieved from
%  http://www.mathworks.com/matlabcentral/fileexchange/22022-matlab2tikz-matlab2tikz
%where you can also make suggestions and rate matlab2tikz.
%
\begin{tikzpicture}

\begin{axis}[%
width=0.956\mywidth,
height=0.987\myheight,
at={(0\mywidth,0\myheight)},
scale only axis,
xlabel style={font=\color{white!15!black}},
xlabel={Number of samples},
xmin=50,
xmax=445,
ylabel style={font=\color{white!15!black}},
ylabel={Macro F1 score},
legend columns=2,
xmajorgrids,
ymajorgrids,
grid style={dotted},ticklabel style={font=\ticklabelfontsize},
legend style={
	at={(0,1.015)}, 
	anchor=south west, legend cell align=left, align=left,
	draw=none
	% white!15!black
	, font=\legendfontsize}
]
\addplot [line width=\mylinewidth,AGNN, mark=*, mark size=\markwidth, mark options={solid}]
table [row sep=\\]{%
55 0.58\\
110 0.60\\
220 0.66\\
440  0.70\\
};
\addlegendentry{TGCN}
\addplot [mark size=\markwidth, line width=\mylinewidth, GCN, dashed, mark=x,  mark options={solid}]
table [row sep=\\]{%
55 0.47\\
110 0.48\\
220 0.48\\
440  0.49 \\
};
\addlegendentry{GCN}
\addplot [mark size=\markwidth, line width=\mylinewidth, AGNNnf, dotted, mark=diamond,  mark options={solid}]
table [row sep=\\]{%
55 0.43\\ 
110 0.41\\
220 0.44\\
440  0.40 \\
};
\addlegendentry{TGCN without features}
\addplot [mark size=\markwidth, line width=\mylinewidth, Mune, dashdotted, mark=square,  mark options={solid}]
table [row sep=\\]{%
55  0.13\\
110 0.24\\
220 0.25\\
440  0.28\\
};
\addlegendentry{Mune}
\end{axis}
\end{tikzpicture}%
    \caption{Generic cells}
    \label{fig:gc}
\end{figure}

Figs. 9-\ref{fig:gc} report the macro F1 values for the aforementioned approaches 
for varying numbers of labeled samples $|\mathcal{M}|$. It is observed for all datasets  that:
i) the macro F1 score improves 
for increasing $|\mathcal{M}|$ across all algorithms; ii) the TGCN that judiciously 
combines the multiple-relations outperforms the GCN by a large margin; and, iii) When nodal features are not used  (last two rows at each table), TGCN outperforms the state-of-the-art Mune.

\section{Conclusions}
This paper put forth a novel deep SSL approach based on a tensor graph convolutional network (TGCN).  The proposed architecture is able to account for nodes engaging in multiple relations, can be used to reveal the data structure, and it is computationally affordable since the number of operations scales linearly with the number of graph edges. Instead of committing a fortiori to a specific type of diffusion, the TGCN learns the diffusion pattern that best fits the data. Our TGCN was also adapted to robustify SSL over a single graph with model-based, adversarial or random edge perturbations. To cope with adversarial perturbations, random edge dithering (ED) was performed on the (nominal) graph edges, and the dithered graphs were used as input to the TGCN. Our approach achieved state-of-the-art classification results over multi-relational graphs when nodes are accompanied by feature vectors. Further experiments demonstrate the performance gains of TGCN in the presence of noisy features, noisy edge weights, and random as well as adversarial edge perturbations. Future research includes predicting time-varying labels, and using TGCN for nonlinear Bayesian estimation over graphs. 

\bibliographystyle{IEEEtran}
\bibliography{my_bibliography}
\noindent

\end{document}